\documentclass[10pt,journal,compsoc]{IEEEtran}

\ifCLASSOPTIONcompsoc
   \usepackage{setspace}
   \usepackage[pdftex]{graphicx}
   \usepackage{amsmath,epsfig,amssymb,subfigure,bm,dsfont}
   \usepackage{multirow}
   \usepackage{epstopdf}
   \usepackage{tabu}
   \usepackage{color}
   \usepackage{balance}
   \usepackage[citecolor=blue, colorlinks]{hyperref}
   \usepackage{soul}
   \usepackage{mathrsfs}
    \usepackage{bbding}
  \usepackage[nocompress]{cite}
\else
  \usepackage{cite}
\fi
\newcommand{\Hquad}{\hspace{0.5em}}

\begin{document}
%
\title{Deep Diversity-Enhanced Feature Representation of Hyperspectral Images}

\author{Jinhui~Hou,~Zhiyu~Zhu,~Junhui~Hou, \textit{Senior Member, IEEE},~Hui~Liu,~Huanqiang Zeng, \textit{Senior Member, IEEE}, and Deyu Meng, \textit{Member, IEEE}
\IEEEcompsocitemizethanks{
\IEEEcompsocthanksitem J. Hou, Z. Zhu, and J. Hou are with the Department of Computer Science, City University of Hong Kong, Hong Kong (e-mail: jhhou3-c@my.cityu.edu.hk; zhiyuzhu2@my.cityu.edu.hk; jh.hou@cityu.edu.hk).
\IEEEcompsocthanksitem H. Liu is with the School of Computing \& Information Sciences, Saint Francis University, Hong Kong. (e-mail: h2liu@sfu.edu.hk).
\IEEEcompsocthanksitem H. Zeng is with the School of Engineering, Huaqiao University, Quanzhou 362021, China, and also with the School of Information Science and Engineering, Huaqiao University, Xiamen 361021, China (e-mail: zeng0043@hqu.edu.cn).

\IEEEcompsocthanksitem D. Meng is with
the School of Mathematics and Statistics, Xi'an Jiaotong University, Xi'an
710049, China (e-mail: dymeng@mail.xjtu.edu.cn).

\IEEEcompsocthanksitem The first two authors contributed to this paper equally. This work was supported in part by the Hong Kong Research Grants Council under Grants 11219019, 11218121, and 11219422, and in part by Hong Kong  Innovation and Technology Fund MHP/117/21. \textit{Corresponding author: Junhui Hou}.
}
}


\IEEEtitleabstractindextext{%
\begin{abstract}
In this paper, we study the problem of efficiently and effectively embedding the high-dimensional spatio-spectral information of hyperspectral (HS) images, guided by feature diversity. Specifically, based on the theoretical formulation that feature diversity is correlated with the rank of the unfolded kernel matrix, we rectify 3D convolution by modifying its topology to enhance the rank upper-bound. This modification yields a rank-enhanced spatial-spectral symmetrical convolution set (ReS$^3$-ConvSet), which not only learns diverse and powerful feature representations but also saves network parameters. Additionally, we also propose a novel diversity-aware regularization (DA-Reg) term that directly acts on the feature maps to maximize independence among elements. To demonstrate the superiority of the proposed ReS$^3$-ConvSet and DA-Reg, we apply them to various HS image processing and analysis tasks, including denoising, spatial super-resolution, and classification. Extensive experiments show that the proposed approaches outperform state-of-the-art methods both quantitatively and qualitatively to a significant extent. The code is publicly available at \url{https://github.com/jinnh/ReSSS-ConvSet}.
\end{abstract}

}

\maketitle

\IEEEdisplaynontitleabstractindextext

%
\IEEEpeerreviewmaketitle

\ifCLASSOPTIONcompsoc
\IEEEraisesectionheading{\section{Introduction}\label{sec:introduction}}
\else
\section{Introduction}
\label{sec:introduction}
\fi

%
%
%
%
\IEEEPARstart{T}{he} advent of hyperspectral (HS) imaging has opened up new vistas in the ability to analyze and understand the complex fabric of real-world scenes and objects. By capturing a vast spectrum of light beyond the capabilities of the human eye, HS imaging has become an indispensable tool in fields as diverse as environmental monitoring \cite{shimoni2019hypersectral}, precision agriculture \cite{park2015hyperspectral}, and marine ecosystem assessment \cite{Zhong2005Absorption}. The unprecedented level of detail offered by HS images, encapsulating both spatial and spectral information, holds the promise for nuanced analysis unattainable by traditional imaging techniques.

Despite the potential, the high-dimensional nature of HS data poses a significant challenge: how to effectively and efficiently extract and embed this rich spatial-spectral information? Deep learning techniques have catalyzed a transformative era of innovation. Pioneering works leveraged various networks and techniques, including convolutional neural networks \cite{Chang2019HSI,Jiang2020Learning,Cao2021Deep}, generative adversarial networks \cite{hang2021classification}, recurrent neural networks \cite{mou2017deep},  transformers \cite{hong2022spectralformer,li2023spatial,zhang2023essaformer}, and stacked kinds of modules like multi-scale strategy \cite{Yuan2019Hyperspectral,he2019feature,Zhang2019Hybrid,wan2022a}, attention mechanism \cite{zhu2021residual,paoletti2022multiple}, and atrous convolution \cite{Liu2019A,Shi2021Hyperspectral}, to enhance feature extraction and model performance. However, these approaches often entail substantial computational demands, complexity, and potential stability issues during the learning process. Furthermore, a recurring issue is effectively balancing the extraction of spatial and spectral information. Some methods \cite{Shi2021Hyperspectral,hong2022spectralformer} overly prioritize one type of information at the expense of the other or fail to account for their interdependencies.

Among the array of techniques, 3D convolution stands out as the most straightforward approach to processing HS images, has been widely employed to simultaneously convolve in spatial and spectral domains \cite{Mei2017Hyperspectral,Zhang2019Hybrid,hamida20183,Yuan2019Hyperspectral,wang2021hyperspectral}.
However, despite its advantages, the use of 3D convolution often comes at the expense of increased computational demand. To mitigate this, pioneering works \cite{qiu2017learning,tran2018closer,gonda2018parallel,Dong2019Deep} delved into optimizing computation through the empirical synthesis of 2D and 1D convolutions targeting orthogonal dimensions. However, these strategies frequently proceed without a comprehensive investigation of network architecture, typically leading to both theoretically and empirically sub-optimal solutions.

\textcolor{black}{Our study diverges from the extant literature by placing a spotlight on this aspect of feature diversity, which has been shown crucial for performance in HS applications \cite{zhang2018diverse,yang2022diversity}. The rationale for this focus stems from the realization that a diverse feature representation is instrumental in encapsulating the full breadth of information present in rich HS imagery. Learning a varied and independent set of features enables capturing subtle yet decisive spectral-spatial distinctions within the data. This capacity is paramount for HS image analysis, where nuanced differences, though finely grained, often carry critical domain-specific significance. In light of this motivation, by starting with 3D convolution and delving into the theoretical aspects of feature representation, we quantify feature diversity using the rank of the 2D matrix formed by convolutional kernels. This novel perspective reframes the problem of learning diverse spatio-spectral representations as the task of increasing the upper bound of this rank.}

To achieve this, we ingeniously modify the network topology, resulting in our rank-enhanced spatial-spectral symmetrical convolution set (ReS$^3$-ConvSet). ReS$^3$-ConvSet enables the learning of diverse spatial-spectral features while also reducing the number of network parameters. Additionally, we propose a novel diversity-aware regularization (DA-Reg) term, which directly optimizes the distribution of the singular values of the 2-D matrix formed by features to promote diversity.
Finally, based on the proposed ReS$^3$-ConvSet and DA-Reg, we construct three learning-based frameworks for HS image denoising, spatial super-resolution, and classification to demonstrate their significance. Extensive experiments on commonly used HS image benchmark datasets demonstrate the significant superiority of the proposed methods over state-of-the-art ones.

In summary, the main contributions of this paper are four-fold:
\begin{itemize}

\item \textcolor{black}{We introduce ReS$^3$-ConvSet, a theoretically grounded spatial-spectral feature representation module that not only efficiently captures diverse feature representations but also simplifies network complexity.}

\item \textcolor{black}{We propose a diversity-aware regularization term, serving as a potent tool to engrain feature diversity within the learning paradigm, ultimately boosting the discriminative capacity of the features.}

\item \textcolor{black}{We develop and validate leading-edge learning-based frameworks for HS image denoising, super-resolution, and classification, demonstrating marked performance improvements over existing state-of-the-art methodologies.}

\item \textcolor{black}{Beyond the proposed techniques themselves, our work provides a unified perspective on designing diverse feature learning modules for HS analysis.}

\end{itemize}

The remainder of the paper is organized as follows. Section~\ref{sec:Re} reviews the existing related works, including HS image denoising, HS image spatial super-resolution, and HS image classification. Section~\ref{sec:proposed} describes our proposed ReS$^3$-ConvSet in detail, which is further incorporated into various HS image processing tasks in Section \ref{sec:application}, including HS image denoising, spatial super-resolution, and classification. In Section~\ref{sec:experiments}, we conduct extensive experiments and comparisons to demonstrate the advantages of the proposed methods,  as well as comprehensive ablation studies. Finally, Section~\ref{sec:con} concludes this paper.

\section{Related Work}
\label{sec:Re}

In this section, we briefly review deep learning-based methods for HS image processing and analysis, including HS image denoising, spatial super-resolution, and classification. We also refer readers to \cite{Ghamisi2017Advances,li2019deep,peng2022low} for the comprehensive survey on these topics.

\vspace{1em} \noindent
\textbf{HS Image Denoising}
In the quest to enhance HS image denoising, deep learning has offered significant advancements over traditional optimization methods. Chang et al. \cite{Chang2019HSI} spearheaded this movement with 2-D deep neural networks; however, the spectral dimension was not well exploited. Yuan et al. \cite{Yuan2019Hyperspectral} responded with a spatial-spectral network that integrates 2-D and 3-D convolutions, but computational efficiency remains a concern.
Then, Liu et al. \cite{Liu2019A} innovated with a 3-D atrous network for wider receptive fields.
Dong et al. \cite{Dong2019Deep} then deconstructed 3-D convolutions into 2-D spatial and 1-D spectral operations, achieving efficiency but risking insufficient feature integration.
Wei et al. \cite{Wei20213D} contributed a 3-D quasi-recurrent network, weaving in structural spatio-spectral and global correlations. However, it requires more complex architectures and training than standard convolutional neural networks.
Zhang et al. \cite{zhang2021lr} integrated low-rank properties to harness inherent data structures, yet it may suffer from high computational complexity.
Shi et al. \cite{Shi2021Hyperspectral} introduced a dual-branch 3-D attention network to dynamically target relevant features, but it does not adequately address the inherent complexities in the correlation between the spatial and spectral information.
Cao et al. \cite{Cao2021Deep} constructed a global reasoning network, merging local and global information, while the balance between local and global information processing was not fully explored.
Rui et al. \cite{Rui2021CVPR} adopted a data-driven approach for a generalizable denoising model, yet its adaptability to diverse noise profiles was untested. Bodrito et al. \cite{Bodrito2021a} combined sparse coding with deep learning to create a trainable model, which, however, may lack broad generalizability.

\vspace{1em} \noindent
\textbf{HS Image Spatial Super-resolution.}
In the realm of HS image spatial super-resolution, deep learning has catalyzed a transformative era of innovation.
Mei et al. \cite{Mei2017Hyperspectral} introduced a 3-D FCNN that navigated the spatial and spectral dimensions simultaneously, marking a step towards unified feature extraction but with substantial computational demands.
Jiang \emph{et al.} \cite{Jiang2020Learning} proposed a spatial-spectral prior framework, called SSPSR, equipped with grouped convolutions and progressive upsampling strategies.
Li et al. \cite{Li2020Hyperspectral} infused generative adversarial networks into the task, enhancing detail through adversarial training, albeit with potential stability issues during the learning process.
Wang et al. \cite{wang2021hyperspectral} pursued a dual-channel approach, blending 2-D and 3-D CNNs for a comprehensive learning schema that, while robust, scaled the complexity of the model. Hu et al. \cite{Hu2020Hyperspectral} proposed an intrafusion network that synergized spectral differences and parallel convolutions, though it may have overlooked the global coherence of spectral bands.
The interplay between 2-D and 3-D convolutions was further scrutinized by Li et al. \cite{Li2021Exploring}, aiming for a balanced fusion of spatial-spectral information but without a clear strategy for prioritizing feature relevance. Multi-task learning was adopted by Li et al. \cite{li2022hyperspectral} to address HS and RGB super-resolution concurrently, a commendable strategy that increased the model's versatility but also its training complexity.
Wang et al. \cite{wang2022hyperspectral} presented a recurrent feedback network that sequentially captured spectral band dependencies, offering a dynamic perspective on spectral information flow but potentially increasing the complexity of the model. Hou et al. \cite{hou2022deep} developed a deep posterior distribution-based embedding framework, extracting features from a probabilistic standpoint, yet it requires higher parallel resources to ensure efficiency.

\vspace{1em} \noindent
\textbf{HS Image Classification.}
Early endeavors in HS image classification, such as those employing support vector machines (SVMs) \cite{melgani2004classification}, random forest (RF) \cite{ham2005investigation}, Gaussian process \cite{bazi2009gaussian}, and K-nearest neighbors (KNNs) \cite{ma2010local}, demonstrated the potential of machine learning in spectral data analysis but lacked the sophistication to fully decipher the spatial-spectral complexity of HS imagery.
The advent of deep learning heralded a new era of possibilities. Hu et al. \cite{hu2015deep} harnessed 1-D CNNs for spectral feature extraction, while Chen et al. \cite{chen2016deep} and Hamida et al. \cite{hamida20183} expanded this to 3-D CNNs, capturing both spectral and spatial information but often at high computational costs.
Paoletti et al. \cite{paoletti2019deep} introduced a pyramidal residual network that bolstered classification accuracy with its hierarchical structure, while suffering from high computation as well.
Mou et al. \cite{mou2017deep} broke convention by treating HS pixels as sequential data in their RNN model, yet it ignores the correlation between spatial and spectral domains.
Hong \emph{et al.} \cite{hong2021graph} introduced a minibatch GCN and investigated CNN and GCN with different fusion approaches, while it failed to take full advantage of the rich spectral information of HS image.
Generative adversarial approaches, as explored by Hang et al. \cite{hang2021classification}, and the SpectralFormer, a transformer-based network by Hong et al. \cite{hong2022spectralformer}, pushed the envelope further in feature extraction. However, the former grappled with the intricacies of adversarial training, and the latter's focus on spectral information occasionally came at the expense of spatial detail.
In addition, deep learning strategies like multi-scale processing \cite{he2019feature,wan2022a}, attention mechanism \cite{zhu2021residual,paoletti2022multiple} were integrated to enhance classification performance, yet the complexity of such models often led to intensive computation.

\vspace{1em} \noindent
\textbf{Discussions.}
In the fields of HS image denoising, spatial super-resolution, and classification, deep learning has made significant strides in feature extraction and computational complexity. However, key challenges remain. While some models innovatively exploit spatial and spectral dimensions, they often struggle to balance these elements, leading to either insufficient exploitation of spectral dimensions or overlooking of global coherence. Moreover, computational efficiency is a recurring concern; many models incur high computational demands due to complex architectures or the integration of advanced strategies such as multi-scale processing, attention mechanisms, or sophisticated convolution operations.
In response to the limitations identified in HS image-based tasks, our proposed method aims to efficiently and effectively embed high-dimensional spatio-spectral information from the perspective of enhancing feature diversity. This approach not only learns diverse and powerful feature representations but also reduces computational demands, providing a more balanced and efficient model for HS image processing tasks.

\section{Proposed Method}
\label{sec:proposed}

Despite the rich information contained in HS images that facilitates subsequent applications, their high-dimensional nature allows for various possibilities, making it difficult to identify the most effective approach.
Previous studies \cite{Chen2017TrainingGO,wang2020orthogonal} have demonstrated that enhancing feature diversity is a promising solution to overcome network redundancy, thus fully utilizing the model capacity. Thus, we approach this fundamental issue by focusing on promoting feature diversity through quantitative modeling, which further guides us in designing the efficient yet powerful feature embedding module.

\begin{figure*}[!t]
\centering
\includegraphics[width=1\linewidth]{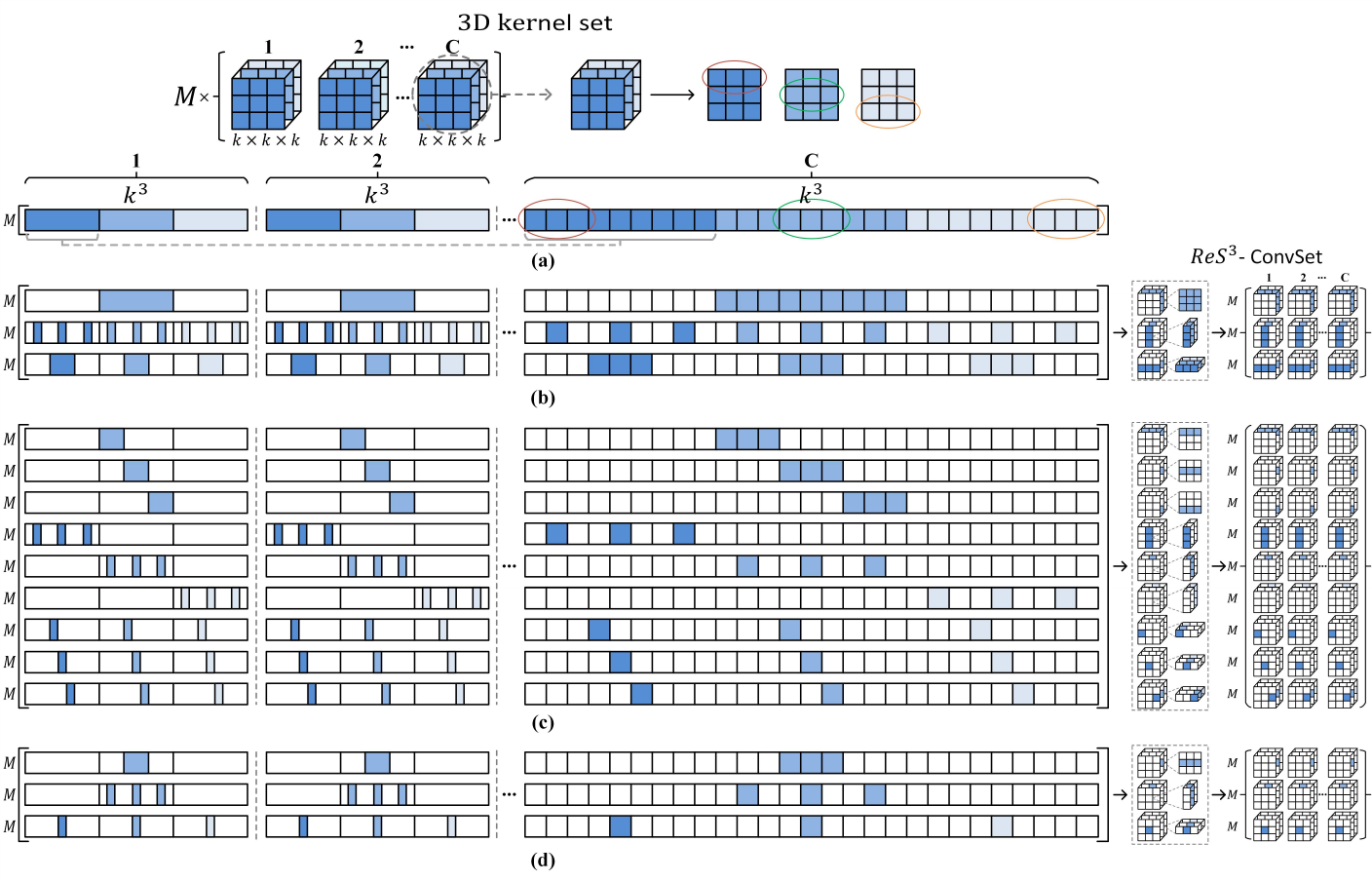}\vspace{-0.35cm}
\caption{Illustration of re-permuting the 3-D convolutional kernel set (a) into rank-enhanced spatial-spectral symmetrical patterns with different rank upper-bounds in (\textbf{b}), (\textbf{c}), and (\textbf{d}). Specifically, The colored cubes symbolize the kernel weights, whereas the blank ones represent replenished zeros. (\textbf{a}) the unrolled 3-D convolutional kernel matrix with the kernel size of 3. Meanwhile, (\textbf{b}), (\textbf{c}), and (\textbf{d}) represent to re-permute 3-D convolutional kernel matrix in our rank-enhanced spatial-spectral symmetrical manner with the kernel size of $3\times 3$, $1\times 3$, and $1\times 3$, of which the scaling factors are $3$, $9$, and  $3$, individually. }
\label{fig:filter-rank}
\end{figure*}

\subsection{Quantitative Modeling of Feature Diversity}
\label{Sec:diversity}

Let $\mathcal{A}\in\mathbb{R}^{M\times C\times k\times k\times k}$ be the kernel tensor of a typical 3-D convolutional layer with $M$ 3-D kernels of size $k\times k\times k$, where $C$ is the number of input feature channels. By feeding an HS feature map $\mathcal{I}\in \mathbb{R}^{C\times B\times H\times W}$, the 3-D convolutional layer outputs feature map 
$\mathcal{F}\in \mathbb{R}^{M\times B'\times H'\times W'}$.
We can equivalently express the 3-D convolution process in the form of 2-D matrix multiplication \cite{heide2015fast,yanai2016efficient}, i.e.,
\begin{equation}
\label{equ:convmatrix}
\mathbf{F} = \mathbf{A} \cdot \mathbf{I},
\end{equation}
where $\mathbf{F} \in \mathbb{R}^{M \times B'H'W'}$ is the matrix form of
$\mathcal{F}$, $\mathbf{A} \in \mathbb{R}^{M \times k^3C}$ is the kernel matrix derived by flattening each of the 3-D kernels as a row vector and then vertically stacking them together (as illustrated in Fig. \ref{fig:filter-rank}(\textcolor{red}{a})), and $\mathbf{I} \in \mathbb{R}^{k^3C \times B'H'W'}$ denotes the matrix form of $\mathcal{I}$ obtained by iteratively unrolling it along the spatial-spectral directions. 

Considering that the rank of a 2-D matrix indicates the independence/freedom of its elements, we analyze the rank of the feature matrix $\mathbf{F}$ to quantitatively measure its diversity. 
Specifically, based on the property of matrix multiplication, from Eq. (\ref{equ:convmatrix}), we have
\begin{equation}
\texttt{Rank}(\mathbf{F}) \leq \texttt{min}\{\texttt{Rank}(\mathbf{A}), \texttt{Rank}(\mathbf{I})\},
\end{equation}
where $\texttt{Rank}(\cdot)$ returns the rank of an input 2-D matrix. As the values of $M$ and $C$ are usually at the same level,
we have $M \ll k^3C \ll B'H'W'$, resulting in
$\texttt{Rank}(\mathbf{F})\leq M$.
Therefore, relieving the upper bound of $\texttt{Rank}(\mathbf{F})$, which is fundamentally limited by that of $\texttt{Rank}(\mathbf{A})$, 
is the prerequisite to promote the learning of diverse and powerful features.
To this end,  we propose \textcolor{black}{a rank-enhanced spatial-spectral symmetrical convolution set (ReS$^3$-ConvSet)} to raise the upper bound of $\texttt{Rank}(\mathbf{A})$ by altering the network topology directly, i.e., structure-level feature diversity enhancement (Section \ref{Sec:ReConv}). 
Besides, considering that a large proportion of kernel weights tend to converge to a marginal set of principal components during back-propagation \cite{denil2013predicting,shang2016understanding,Lin2020HRank}, 
we further explicitly regularize the feature matrix $\mathbf{F}$ to make it deviate from an approximately low-rank one, i.e., content-level feature diversity enhancement (Section \ref{Sec:Rankloss}).

\subsection{Structure-level Feature Diversity Enhancement} %
\label{Sec:ReConv}

As shown in Fig. \ref{fig:filter-rank} (\textcolor{red}{a}), the convolutional kernels contained in a typical 3-D convolutional layer is flattened into a fat 2-D matrix 
$\mathbf{A} \in \mathbb{R}^{M \times k^3C}$, 
the rank of which is upper bounded by $M$.
Generally, we relieve such a limitation by permuting the elements of $\mathbf{A}$ into a larger 2-D matrix with a higher rank upper bound, which can be further realized by a typical convolution manner.
Note that as the three dimensions of HS images are equally essential \cite{Liu2019A,zhang2016simultaneous}, we process the three dimensions indiscriminately to ensure that an identical convolution pattern could be finally applied to the three dimensions (i.e., the symmetry property).  

Technically, we first construct a larger kernel matrix $\mathbf{A}_{a} \in \mathbb{R}^{3LM \times k^3C}$ by scaling up the number of channels by $3L$ ($L$ is a positive integer), i.e., each of the three dimensions is equipped with $LM$ channels. Accordingly, we can rewrite the convolution process as
\begin{equation}
\begin{split}
\label{equ:rankenchance-conv}
\mathbf{F}_{a} = \mathbf{A}_{a} \cdot \mathbf{I}~~~{\rm with}~~
\mathbf{A}_{a} = [\mathbf{A}_{a}^{1};\mathbf{A}_{a}^{2};\mathbf{A}_{a}^{3}], 
\end{split}
\end{equation}
where $\mathbf{F}_{a}\in \mathbb{R}^{3LM \times B'H'W'}$ denotes the resulting feature volume, and $\{\mathbf{A}_{a}^{i}\in \mathbb{R}^{LM \times k^3C}\}_{i=1}^3$ are the kernel matrices corresponding to the three dimensions of HS images.
We expect that $\mathbf{A}_a$ corresponds to a typical convolution manner with the number of parameters not exceeding that of 3D convolution (i.e., the number of elements in $\mathbf{A}$), which requires that $\mathbf{A}_a$ has to be replenished with a certain number of zeros. Moreover, we obtain that the maximum number of parameters involved in each convolutional kernel is $\frac{k^3C}{3L}$.
Besides, we expect that the resulting kernel matrix $\mathbf{A}_a$ can be easily realized by commonly used convolution patterns. More specifically, we select the widely disseminated \textit{regular} 1-D/2-D convolution as the desired alternative.
In the following we take the case that $k=3$ ($k^3=27$ and $\frac{k^3C}{3L}=\frac{9C}{L}$) as an example for convenience. A larger convolutional kernel could be approximated by sequentially aggregating multiple layers with small kernels.

\begin{figure*}[t]
\centering
\includegraphics[width=0.9\linewidth]{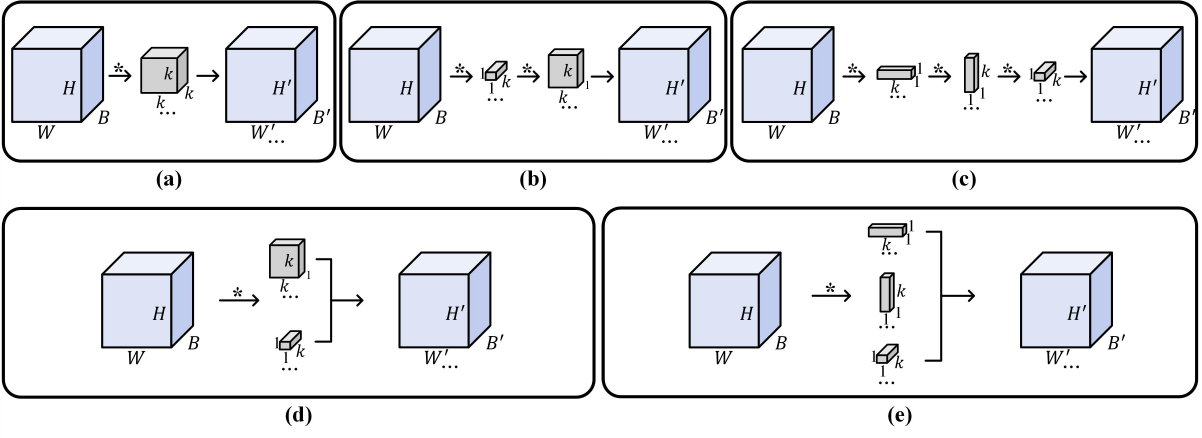}\vspace{-0.35cm}
\caption{Illustration of various feature extraction manners for HS images.
(a) 3-D convolution, (b) Sequential 1-D and 2-D convolution, (c) Sequential 1-D convolution, (d) 1-D + 2-D convolution, and (e) Proposed ReS$^3$-ConvSet.}
\label{fig:low-rank-filters}
\end{figure*}

Based on the above conditions, when $L=1$, we can deduce one feasible $\mathbf{A}_a$ 
with the rank upper bound \textit{minimally} raised to $3M$, as depicted in Fig. \ref{fig:filter-rank} (\textcolor{red}{b}), where the colored blocks indicate the distribution of network parameters. And such a kernel matrix can be realized by separately applying 2-D convolution of size $3 \times 3$ along the three dimensions.
When $L=3$, as illustrated in Fig. \ref{fig:filter-rank}  (\textcolor{red}{c}), we can construct a feasible $\mathbf{A}_a$ with the rank upper bound \textit{maximally} boosted to $9M$, which corresponds to the process of performing three 1-D convolution of size $3$ in each of the three dimensions.
For the above two feasible solutions, the number of kernel weights in $\mathbf{A}_a$ is equal to that in $\mathbf{A}$. Alternatively, we can also construct feasible $\mathbf{A}_a$ with a higher rank upper bound but fewer parameters to save computational costs. Specifically, as shown in Fig. \ref{fig:filter-rank}  (\textcolor{red}{d}), when $L=1$ (resp. $L=2$), we can build $\mathbf{A}_a$ with the rank upper bound equal to $3M$ (resp. $6M$), which can be achieved by separately applying a single (resp. two) 1-D convolution of size $3$ to each dimension.

Based on the above analyses, we can construct $\mathbf{A}_a$, whose rank upper bound varies from $3M$ to $9M$ without introducing extra parameters, thus potentially promoting the diversity of $\mathbf{F}_a$.
Moreover, in light of the observation that the convolutional kernels for realizing $\mathbf{A}_a$ with the rank upper bound equal to $6M$ or $9M$
share a similar pattern and receptive field to that for the case in Fig. \ref{fig:filter-rank} (\textcolor{red}{d}), they may only result in homogeneous feature content and a minor increase in feature diversity, which is also experimentally demonstrated, i.e., a larger rank upper bound than $3M$ only brings marginal performance improvement (see Table \ref{tab:diff-rub}).
Thus, we finally adopt the permutation manner in Fig. \ref{fig:filter-rank} (\textcolor{red}{d}), i.e., separately performing 1-D convolution along the three dimensions, which can not only learn diverse spatial-spectral feature representations but also save network parameters. 

\noindent
\textbf{Discussions.}
Several previous studies \cite{qiu2017learning,tran2018closer,gonda2018parallel} have also attempted to decompose high-dimensional kernels into a collection of low-dimensional ones to achieve efficient and effective feature embedding. However, considering the different possible compositions (e.g., 1D + 2D, 2D + 2D, or 1D + 1D + 1D) and aggregation methods (e.g., sequential or parallel), without conducting an in-depth investigation of the network structure, this usually results in sub-optimal solutions both theoretically and empirically. In contrast, our work proposes an analytical solution for structuring convolution patterns from the perspective of feature diversity, which is quantified by the rank of feature maps and is positively correlated with the network's learning capacity.
This theoretical formulation guides us to increase the upper bounds of feature diversity by fully decoupling the three dimensions, leading to the utmost effectiveness and efficiency. Importantly, this approach represents unexplored territory not investigated in previous works. Furthermore, the value of our work extends beyond the proposed method itself. Our theoretical framework provides a unified explanation for the various empirically designed feature embedding modules used for volumetric data, thereby advancing the research field in a comprehensive manner.

\begin{figure*}[t]
\centering
\includegraphics[width=0.9\linewidth]{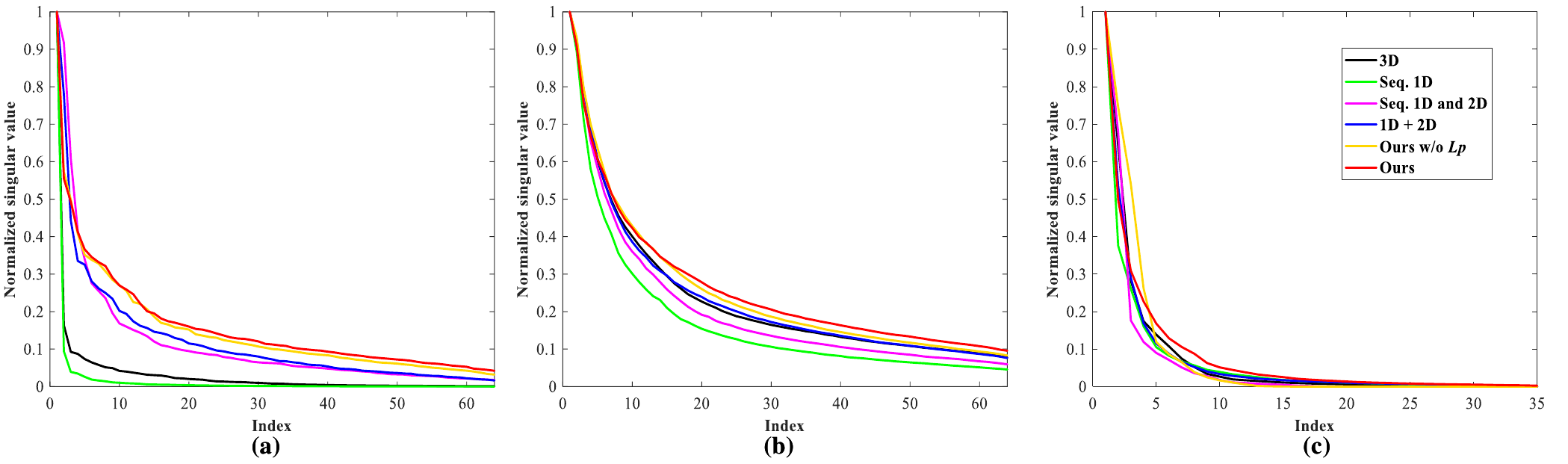}
\vspace{-0.5cm}
\caption{Comparison of the distribution of the singular values of the feature maps extracted by different convolution manners used in the applications of (\textbf{a}) HS image denoising on the ICVL dataset with the Gaussian noise ($\sigma=70$), (\textbf{b}) HS image $4\times$ spatial super-resolution on the CAVE dataset, and (\textbf{c})  HS image classification on  Indian Pines. The singular values of each convolution scheme are normalized by its largest value. 
}
\label{fig:d-sr-cls-ab-sv-comparison}
\end{figure*}

\subsection{Content-level Feature Diversity Enhancement}
\label{Sec:Rankloss}

Although the upper-bound of $\texttt{Rank}(\mathbf{F})$  has been increased through kernel matrix re-permutation, there remains the theoretical potential that the resulting feature diversity, represented by $\mathbf{F}_a$, may not be fully enhanced. Previous studies \cite{denil2013predicting,shang2016understanding,Lin2020HRank} have demonstrated that during back-propagation, a significant portion of kernel weights tends to converge towards a marginal set of principal components. As a result, even if the singular values of a  2-D feature matrix have been boosted from $\{a_i\}_{i=1}^{d}$ to $\{b_i\}_{i=1}^{nd}$, a scenario may arise where a long-tail distribution of singular values occurs: $b_i \approx a_i \Hquad (i\in [1,~d])$, and $ a_i \gg b_j > 0 \Hquad (i\in [1,~d], j\in [d+1,~nd])$. In this case, the additional singular values $\{b_j\}_{j=d+1}^{nd}$  have marginal magnitudes that do not bring notable benefits, despite the re-permuted feature matrix having a significantly higher rank.
In other words, the resulting $\mathbf{F}_a$ may be approximately low-rank. In view of this issue, we further regularize the feature matrix $\mathbf{F}_a$ directly to enhance its diversity and propose the following loss term:
\begin{equation}
\label{equ:fecloss}
\mathcal{L}_{p} = - \|~\texttt{SVD}(\mathbf{F}_a)~\|_1,
\end{equation}
where $\texttt{SVD}(\cdot)$ returns a vector of singular values of an input 2-D matrix, and $\|\cdot\|_1$ is the $\ell_1$ norm.

The proposed regularization term serves to reduce the gaps between large singular values and small ones, promoting a uniform distribution. Accordingly, $\mathbf{F}_a$ is prevented from approximating a low-rank matrix, thereby increasing the freedom (or diversity) of its elements.
It is important to note that we apply this regularization term solely at the last layer of the network, just before the final output. This placement ensures that we avoid introducing excessively high computational costs while still benefiting from the regularization's effects.

\subsection{More Analysis}
\label{sec:discussion}

We also analyze existing feature extraction methods for HS images by using our formulation to get a thorough understanding of them from the perspective of feature diversity, including ``Sequential 1-D and 2-D convolution" in Fig. \ref{fig:low-rank-filters} (\textcolor{red}{b}), ``Sequential 1-D convolution" in Fig. \ref{fig:low-rank-filters} (\textcolor{red}{c}), and ``1-D + 2-D convolution" in Fig. \ref{fig:low-rank-filters} (\textcolor{red}{d}).

\textbf{1}) We can write the process of ``Sequential 1-D and 2-D convolution" in the form of matrix multiplication:
\begin{equation}
\begin{split}
\label{equ:seq1d2d}
\mathbf{F}_{s1d2d} = \mathbf{A}_{2d} \cdot \mathbf{I}_{1d}, ~~
\mathbf{F}_{1d} = \mathbf{A}_{1d} \cdot \mathbf{I},
\end{split}
\end{equation}
where $\mathbf{F}_{s1d2d}\in \mathbb{R}^{M \times B'H'W'}$ is the 2-D matrix form of the output feature maps;
$\mathbf{F}_{1d} \in \mathbb{R}^{M \times B'H'W'}$ denotes the matrix form of the intermediate feature maps by 1-D convolution; $\mathbf{I}_{1d} \in \mathbb{R}^{k^3M \times B'H'W'}$ symbolizes the local patches via sliding 2-D kernels over $\mathbf{F}_{1d}$; and $\mathbf{A}_{1d}$ and $\mathbf{A}_{2d} \in \mathbb{R}^{M \times k^3M}$ are the matrices of 1-D and 2-D kernels, respectively.

\textbf{2}) The convolution process of ``Sequential 1-D convolution" can be phrased as
\begin{equation}
\begin{split}
\label{equ:seq1d}
\mathbf{F}_{s1d} = \mathbf{A}_{1d_3} \cdot \mathbf{I}_{1d_2}, ~
\mathbf{F}_{1d_2} = \mathbf{A}_{1d_2} \cdot \mathbf{I}_{1d_1}, ~
\mathbf{F}_{1d_1} = \mathbf{A}_{1d_1} \cdot \mathbf{I},
\end{split}
\end{equation}
Here, $\mathbf{F}_{1d_1}, \mathbf{F}_{1d_2}, \mathbf{F}_{s1d}\in \mathbb{R}^{M \times B'H'W'}$ indicate the output feature maps of three sequentially aggregated convolutional layers, $\mathbf{I}_{1d_1}, \mathbf{I}_{1d_2} \in \mathbb{R}^{k^3M \times B'H'W'}$ represent local patch of corresponding feature maps in matrix forms, $\mathbf{A}_{1d_1}, \mathbf{A}_{1d_2}, \mathbf{A}_{1d_3} \in \mathbb{R}^{M \times k^3M}$ are the matrices of corresponding 1-D kernels, respectively.

\textbf{3}) The convolution process of ``1-D + 2-D convolution" can be written as
\begin{equation}
\begin{split}
\label{equ:1d2d}
\mathbf{F}_{1d2d} = \mathbf{A}_{1d2d} \cdot \mathbf{I}, ~~~{\rm with}~~
\mathbf{A}_{1d2d} = [\mathbf{A}_{2d};\mathbf{A}_{1d}],
\end{split}
\end{equation}
where $\mathbf{F}_{1d2d}\in \mathbb{R}^{2M \times B'H'W'}$ encodes the output feature maps;
$\mathbf{A}_{1d2d}\in \mathbb{R}^{2M \times k^3M}$ denotes the  matrix of combined 1-D and 2-D kernels. 

According to Eqs. (\ref{equ:seq1d2d}), (\ref{equ:seq1d}) and (\ref{equ:1d2d}),
we have $\texttt{Rank}(\mathbf{F}_{s1d2d})\leq M$, $\texttt{Rank}(\mathbf{F}_{s1d})\leq M$, and $\texttt{Rank}(\mathbf{F}_{1d2d})\leq 2M$. That is, these feature extraction manners expand the rank upper bound from $M$ to $2M$ at most, which is still limited. We refer the readers to Tables \ref{tab:d-ab-comb-icvl}, \ref{tab:sr-ab-comb-cave}, and \ref{tab:cls-ab-comb-ip} for the quantitative results of all the variants illustrated in Fig. \ref{fig:low-rank-filters}.\\

\noindent\textbf{Remarks.}
In the case of feature extraction involving parallel branches, e.g., Figs. \ref{fig:low-rank-filters} (\textcolor{red}{d}) and (\textcolor{red}{e}), we utilize a $1\times 1\times 1$ convolutional layer to compress the multiple output feature volumes before feeding it into the subsequent layer. This compression step is crucial to prevent channel explosion.  Consequently, the feature volumes obtained through different convolutional methods ultimately have the same size, ensuring that the rank upper bounds of the resulting feature matrices are equal.

However, \textcolor{black}{the proposed} ReS$^3$-ConvSet is capable of increasing the rank upper bound of feature matrix $\mathbf{F}$ from $M$ to $3M$ during the feature extraction process. This enhancement has the potential to promote greater diversity among the extracted features.
We refer the readers to Fig. \ref{fig:d-sr-cls-ab-sv-comparison} for the comparison of singular value distributions of the feature maps extracted by various convolution manners, where it is evident that the singular values of the feature matrix by our ReS$^3$-ConvSet decrease at a slower rate compared to other schemes. This observation indicates that our ReS$^3$-ConvSet
achieves a better balance among the singular values, preventing a scenario where only a few large ones dominate the feature space. Consequently, the feature matrix exhibits a higher degree of freedom, avoiding the limitations associated with an approximately low-rank structure and promoting feature diversity.

\section{Applications}
\label{sec:application}
In this section, to demonstrate the advantages of the proposed ReS$^3$-ConvSet and DA-Reg, we construct three learning-based methods for HS image denoising, spatial super-resolution, and classification. Note that we mainly incorporate ReS$^3$-ConvSet into existing frameworks to directly demonstrate their advantages.
\begin{figure}[!t]
\centering
\includegraphics[width=1\linewidth]{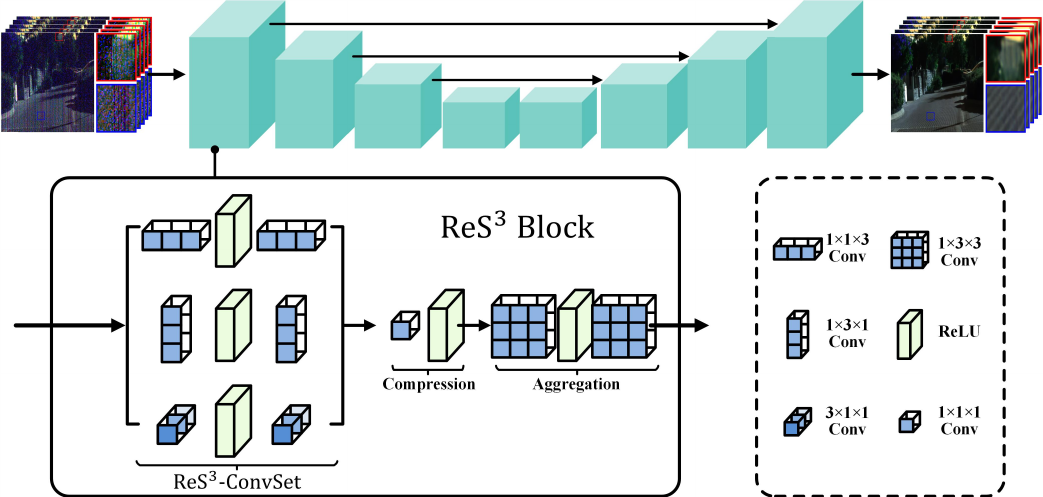}
\caption{Illustration of our HS image denoising framework, which is constructed by incorporating the proposed ReS$^3$-ConvSet into a residual U-Net architecture.
}
\label{fig:hsid-framework}
\end{figure}

\begin{figure}[!t]
\centering
\includegraphics[width=1\linewidth]{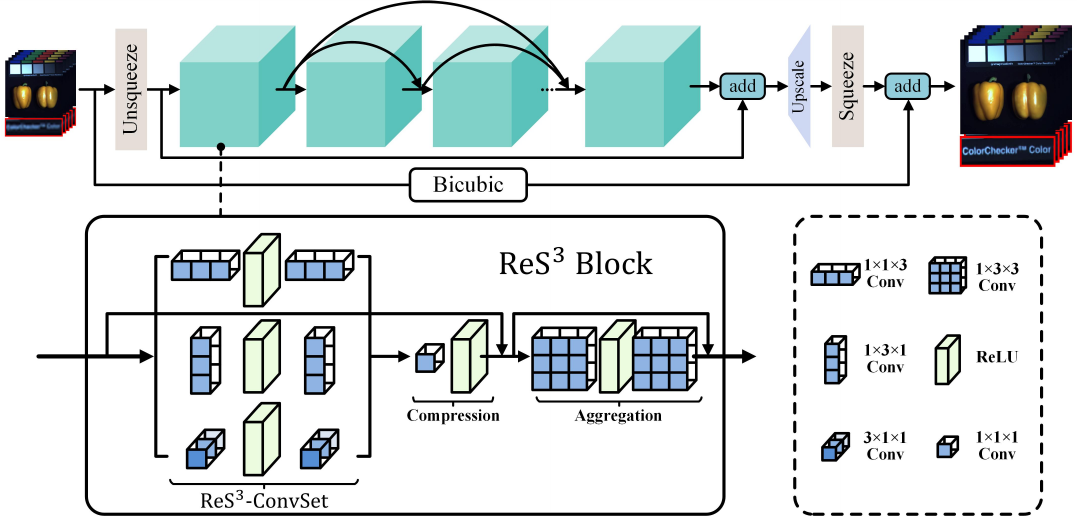}\vspace{-0.35cm}
\caption{Illustration of our HS image super-resolution framework, which is constructed by incorporating the proposed ReS$^3$-ConvSet into a residual-dense architecture.
}
\label{fig:hsisr-framework}
\end{figure}

\subsection{HS Image Denoising}

Let $\mathcal{X}_{n}\in\mathbb{R}^{B\times H\times W}$ be a noisy HS image, and $\mathcal{Y}\in\mathbb{R}^{B\times H\times W}$ the corresponding noise-free one, where $H$ and $W$ are the spatial dimensions, and $B$ is the number of spectral bands. The degradation process of $\mathcal{Y}$ to $\mathcal{X}_{n}$ could be generally formulated as
\begin{equation}
\label{equ:1}
\mathcal{X}_{n} = \mathcal{Y} + \mathcal{N}_{z},
\end{equation}
where $\mathcal{N}_{z}\in \mathbb{R}^{B\times H \times W}$ denotes the additive noise.

We incorporate the proposed ReS$^3$-ConvSet into the widely-used U-Net \cite{ronneberger2015u} architecture for constructing an efficient and compact HS image denoising method.
As illustrated in Fig.~\ref{fig:hsid-framework}, the denoising framework is composed of multiple ReS$^3$ blocks. In each block, ReS$^3$-ConvSet separately performs 1-D convolution along the three dimensions of an HS image, and a compression layer and an aggregation layer are adopted to aggregate the resulting spatial-spectral embeddings.

We train the denoising network by jointly minimizing the $\ell_1$ distance between the recovered HS image $\mathcal{\widehat{Y}}_{d}$ and
$\mathcal{Y}$ and the proposed DA-Reg: 
\begin{equation}
\label{equ:dnoise-lossoverall}
\mathcal{L}_{d} = \mathcal{L}_1(\mathcal{\widehat{Y}}_{d}, \mathcal{Y}) + \lambda \mathcal{L}_{p}.
\end{equation}

\subsection{HS Image Spatial Super-resolution}

Given a low-spatial-resolution HS image denoted as $\mathcal{X}_{lr}$ $\in\mathbb{R}^{B\times h\times w}$ with $h \times w$ being the spatial dimensions, we aim to recover a high-spatial-resolution HS image denoted as $\mathcal{Y}$ $\in\mathbb{R}^{B\times H\times W}$, in which $H=\alpha h$ and $W=\alpha w$ ($\alpha>1$ is the scale factor).
The degradation process of $\mathcal{X}_{lr}$ from $\mathcal{Y}$ can be generally formulated as
\begin{equation}
\label{equ:sr_degradation}
\mathcal{X}_{lr} = \mathscr{D}(\mathcal{Y}) + \mathcal{N}_{z},
\end{equation}
where $\mathscr{D}(\cdot)$ 
is the degeneration operator, 
consisting of the down-sampling and blurring operations.

For HS image spatial super-resolution, as shown in Fig.~\ref{fig:hsisr-framework}, we integrate the proposed method into the widely adopted residual dense architecture \cite{zhang2018residual} to construct an HS image super-resolution framework comprising multiple ReS$^3$ blocks. Each block is devised utilizing residual learning, containing a ReS$^3$-ConvSet, a compression layer, and an aggregation layer. We train the super-resolution framework by simultaneously minimizing the $\ell_1$ distance between the super-resolved HS image $\mathcal{\widehat{Y}}_{sr}$ and $\mathcal{Y}$, and the proposed DA-Reg. The overall loss function is written as
\begin{equation}
\label{equ:sr-lossoverall}
\mathcal{L}_{sr} = \mathcal{L}_1(\mathcal{\widehat{Y}}_{sr}, \mathcal{Y}) + \lambda \mathcal{L}_{p}.
\end{equation}
\begin{figure}[!t]
\centering
\includegraphics[width=1\linewidth]{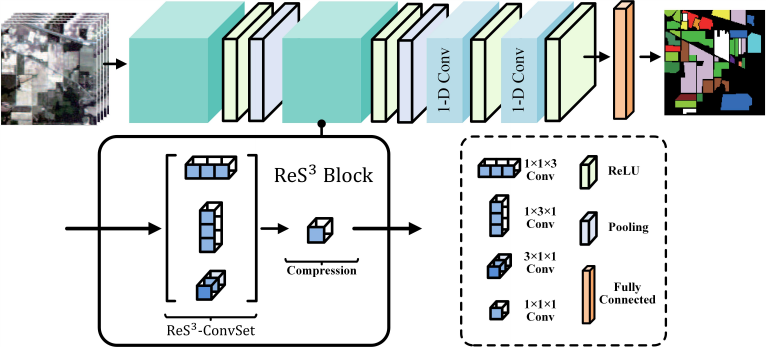}\vspace{-0.35cm}
\caption{Illustration of our HS image classification framework, which is constructed by incorporating the proposed ReS$^3$-ConvSet into the existing method \cite{hamida20183}.
}
\label{fig:cls-framework}
\end{figure}

\subsection{HS Image Classification}

For HS image classification, we incorporate the proposed method into the existing method \cite{hamida20183} to construct the classification framework, as shown in Fig.~\ref{fig:cls-framework},
it consists of one ReS$^3$ block and three conventional filters, i.e., a 3-D convolution for obtaining the basic HS features, and two 1-D convolutions for pooling in the spectral domain.
We train the classification network by simultaneously optimizing the cross entropy loss between predicted labels and ground-truth ones and our DA-Reg, i.e.,
\begin{equation}
\label{equ:classification-lossoverall}
\mathcal{L}_{c} = \mathcal{L}_{ce} + \lambda \mathcal{L}_{p}.
\end{equation}

\section{Experiments}
\label{sec:experiments}

\begin{table*}[t]
\caption{
Quantitative comparisons of different methods under several noise levels over the \textbf{ICVL} dataset. The best and second-best results are highlighted in bold and underlined, respectively. ``$\uparrow$" (resp. ``$\downarrow$") means the larger (resp. smaller), the better.
}\vspace{-0.35cm}
\centering
\begin{spacing}{1.05}
\label{tab:icvlgaussresults}
\resizebox{1\textwidth}{!}{
\begin{tabu}{c|ccc|ccc|ccc|ccc}
\tabucline[1.5pt]{*}
\hline
\multirow{2}{*}{Method}
&\multicolumn{3}{c|}{30} &\multicolumn{3}{c|}{50} &\multicolumn{3}{c|}{70} &\multicolumn{3}{c}{Blind (30-70)}   \\
&MPSNR$\uparrow$ &MSSIM$\uparrow$ &~SAM$\downarrow$~
&MPSNR$\uparrow$ &MSSIM$\uparrow$ &~SAM$\downarrow$~
&MPSNR$\uparrow$ &MSSIM$\uparrow$ &~SAM$\downarrow$~
&MPSNR$\uparrow$ &MSSIM$\uparrow$ &~SAM$\downarrow$~ \\  \hline
Noisy
&18.59 &0.1034 &0.7269
&14.15 &0.0429 &0.9096
&11.23 &0.0228 &1.0273
&14.83 &0.0534 &0.8800 \\
LRMR \cite{Zhang2014Hyperspectral}
&32.96 &0.7247 &0.2298
&28.98 &0.5400 &0.3223
&26.38 &0.4160 &0.3955
&29.57 &0.5674 &0.3060 \\
NMoG \cite{Chen2018Denoising}
&33.24 &0.7464 &0.1537
&29.98 &0.5900 &0.1895
&27.80 &0.4836 &0.2200
&30.48 &0.6146 &0.1823 \\
LRTDTV \cite{Wang2018Hyperspectral}
&38.43 &0.9365 &0.0835
&36.47 &0.9159 &0.1043
&34.97 &0.8971 &0.1201
&36.76 &0.9190 &0.1012 \\
ITSReg \cite{Xie2016Multispectral}
&41.53 &0.9571 &0.0929
&39.19 &0.9350 &0.0841
&37.48 &0.9192 &0.1144
&39.52 &0.9389 &0.1037 \\
GRN \cite{Cao2021Deep}
&41.83 &0.9653 &0.0541
&38.84 &0.9422 &0.1010
&37.22 &0.9264 &0.0853
&39.21 &0.9452 &0.0717 \\
QRNN3D \cite{Wei20213D}
&42.28 &0.9701 &0.0617
&40.22 &0.9544 &0.0733
&38.29 &0.9326 &0.0943
&40.48 &0.9559 &0.0737 \\
NG-Meet \cite{he2022non}
&43.09 &0.9709 &0.0491
&40.35 &0.9541 &0.0582
&38.71 &0.9414 &0.0639
&40.89 &0.9579 &0.0568 \\
T3SC \cite{Bodrito2021a}
&43.19 &0.9718 &0.0616
&40.81 &0.9567 &0.0720
&39.27 &0.9431 &0.0810
&40.98 &0.9579 &0.0723 \\
SST \cite{li2023spatial}
&43.21 &0.9747 &0.0502
&41.10 &0.9611 &0.0585
&39.65 &0.9487 &0.0649
&41.41 &0.9629 &0.0580  \\
SERT \cite{li2023spectral}
&\underline{43.38} &\textbf{0.9752} &\underline{0.0466}
&\underline{41.29} &\textbf{0.9622} &\underline{0.0536}
&\underline{39.84} &\textbf{0.9501} &\underline{0.0593}
&\underline{41.60} &\textbf{0.9640} &\underline{0.0530}  \\
ReS$^3$Net (Ours)
&\textbf{43.66} &\underline{0.9750} &\textbf{0.0443}
&\textbf{41.49} &\underline{0.9614} &\textbf{0.0510}
&\textbf{40.01} &\underline{0.9492} &\textbf{0.0582}
&\textbf{41.80} &\underline{0.9633} &\textbf{0.0513}  \\
\tabucline[1.5pt]{*}
\end{tabu}}
\end{spacing}
\end{table*}

\begin{table*}[t]
\caption{
Quantitative comparisons of different methods under five complex noise cases over the \textbf{ICVL} dataset.
The best and second-best results are highlighted in bold and underlined, respectively. ``$\uparrow$" (resp. ``$\downarrow$") means the larger (resp. smaller), the better.} \vspace{-0.35cm}
\centering
\begin{spacing}{1.2}
\label{tab:icvlcomplexresults}
\resizebox{1\textwidth}{!}{
\begin{tabu}{c|ccc|ccc|ccc|ccc|ccc}
\tabucline[1.5pt]{*}
\hline
\multirow{2}{*}{Method}
&\multicolumn{3}{c|}{1. Non-i.i.d Gaussian} &\multicolumn{3}{c|}{2. Gaussian + Stripe Noise} &\multicolumn{3}{c|}{3. Gaussian + Deadline Noise} &\multicolumn{3}{c|}{4. Gaussian +  Impulse Noise} &\multicolumn{3}{c}{5. Mixture Noise}   \\
&MPSNR$\uparrow$ &MSSIM$\uparrow$ &~SAM$\downarrow$~
&MPSNR$\uparrow$ &MSSIM$\uparrow$ &~SAM$\downarrow$~
&MPSNR$\uparrow$ &MSSIM$\uparrow$ &~SAM$\downarrow$~
&MPSNR$\uparrow$ &MSSIM$\uparrow$ &~SAM$\downarrow$~
&MPSNR$\uparrow$ &MSSIM$\uparrow$ &~SAM$\downarrow$~ \\  \hline
Noisy
&17.80 &0.1516 &0.7911
&17.77 &0.1545 &0.7895
&17.36 &0.1473 &0.8109
&14.86 &0.1118 &0.8480
&14.07 &0.0936 &0.8587  \\
NG-Meet \cite{he2022non}
&28.62 &0.5764 &0.4301
&28.15 &0.5576 &0.4401
&27.69 &0.5703 &0.4443
&25.32 &0.5211 &0.5814
&24.33 &0.4964 &0.5874  \\
LRMR \cite{Zhang2014Hyperspectral}
&28.64 &0.5153 &0.3235
&28.52 &0.5155 &0.3250
&27.78 &0.5075 &0.3398
&24.19 &0.3805 &0.4681
&23.79 &0.3817 &0.4668  \\
NMoG \cite{Chen2018Denoising}
&34.96 &0.8279 &0.1260
&34.60 &0.8184 &0.1793
&33.60 &0.8212 &0.1885
&29.09 &0.6751 &0.4510
&28.45 &0.6746 &0.4568  \\
ITSReg \cite{Xie2016Multispectral}
&36.02 &0.8451 &0.1408
&35.35 &0.8210 &0.1525
&32.68 &0.7879 &0.1812
&26.47 &0.5019 &0.4818
&25.02 &0.4871 &0.4830  \\
LRTDTV \cite{Wang2018Hyperspectral}
&37.95 &0.9377 &0.0671
&37.65 &0.9348 &0.0731
&35.67 &0.9181 &0.0937
&36.60 &0.9265 &0.0874
&34.51 &0.9076 &0.1063  \\
GRN \cite{Cao2021Deep}
&39.97 &0.9587 &0.0685
&39.90 &0.9598 &0.0672
&38.74 &0.9548 &0.0702
&37.63 &0.9410 &0.0952
&38.01 &0.9473 &0.0904  \\
QRNN3D \cite{Wei20213D}
&42.79 &0.9752 &0.0430
&42.64 &0.9750 &0.0437
&42.31 &0.9735 &0.0455
&40.49 &0.9533 &0.0762
&\underline{39.42} &0.9448 &0.0809  \\
T3SC \cite{Bodrito2021a}
&43.51 &0.9776 &0.0441
&43.20 &0.9770 &0.0488
&41.42 &0.9724 &0.0639
&37.93 &0.9353 &0.1669
&35.84 &0.9248 &0.1804  \\
SST \cite{li2023spatial}
&43.45 &\underline{0.9792} &0.0355
&43.29 &0.9789 &0.0364
&42.91 &0.9779 &0.0379
&40.86 &\underline{0.9639} &\underline{0.0573}
&38.62 &0.9470 &\underline{0.0641}  \\
SERT \cite{li2023spectral}
&\underline{43.94} &\textbf{0.9800} &\underline{0.0337}
&\underline{43.89} &\underline{0.9799} &\underline{0.0347}
&\underline{43.56} &\underline{0.9794} &\underline{0.0358}
&\underline{41.08} &0.9602 &0.0643
&39.20 &\underline{0.9487} &0.0734  \\
ReS$^3$Net (Ours)
&\textbf{44.33} &\textbf{0.9800} &\textbf{0.0327}
&\textbf{44.35} &\textbf{0.9805} &\textbf{0.0331}
&\textbf{44.33} &\textbf{0.9802} &\textbf{0.0330}
&\textbf{42.59} &\textbf{0.9668} &\textbf{0.0595}
&\textbf{41.94} &\textbf{0.9664} &\textbf{0.0622}  \\
\tabucline[1.5pt]{*}
\end{tabu}}
\end{spacing}
\end{table*}

\begin{figure*}[!t]
\centering
\includegraphics[width=1\linewidth]{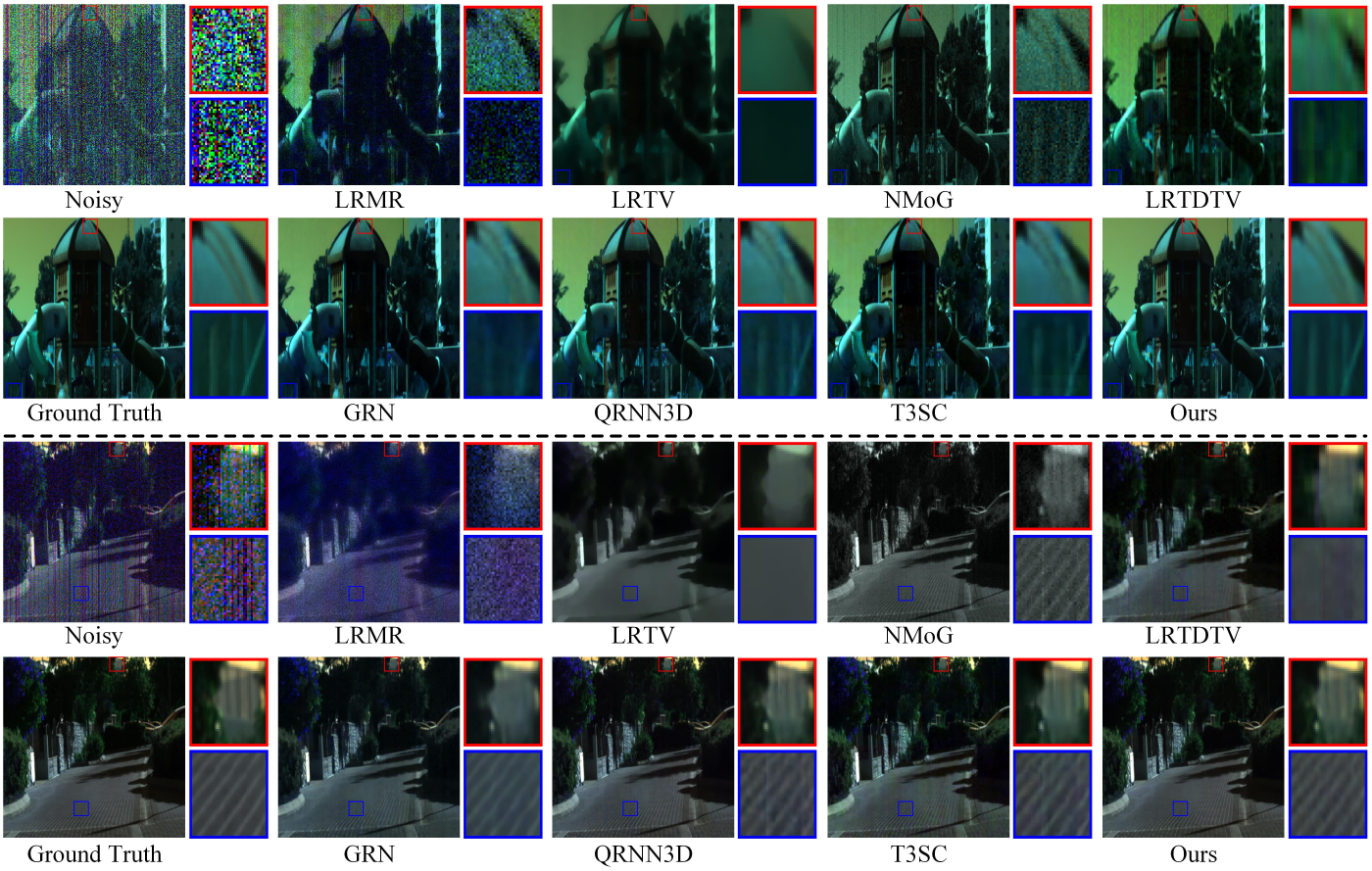}\vspace{-0.35cm}
\caption{Visual comparison of different methods on \textit{ICVL} with Gaussian and deadline, and mixture noise, from top to bottom. Here, we selected the $5^{th}$, $16^{th}$,
and $30^{th}$, and $10^{th}$, $17^{th}$, and $30^{th}$) bands to form a pseudo-RGB image to enable the visualization under Gaussian and deadline noise (resp. mixture noise).
}
\label{fig:icvl-complex}
\end{figure*}

\begin{figure*}[!t]
\centering
\includegraphics[width=1\linewidth]{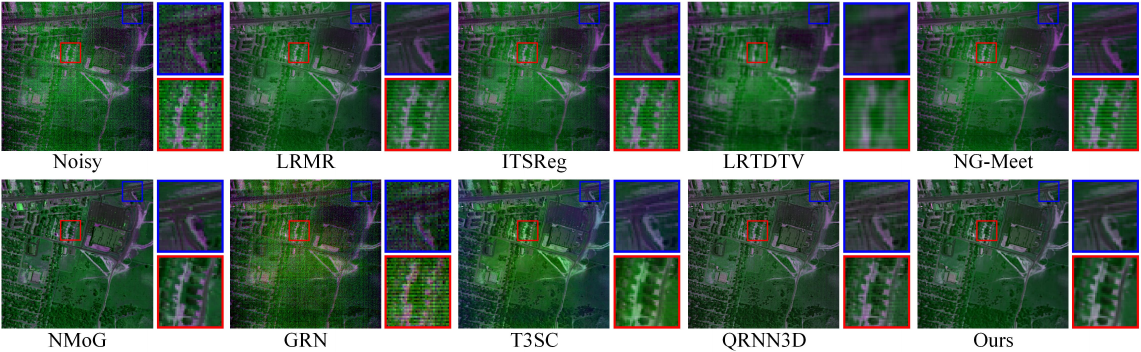}\vspace{-0.35cm}
\caption{Visual comparison of different methods on \textit{Urban} with real unknown noise. Here we selected the $38^{th}$, $112^{nd}$, and $128^{th}$ bands to form a pseudo-RGB image to enable the visualization.
}
\label{fig:urban}
\end{figure*}

\subsection{Evaluation on HS Image Denoising}

\subsubsection{Experiment settings}

\textbf{Datasets.}
We employed two commonly-used HS image benchmark datasets for evaluation, including one natural HS image dataset, i.e., ICVL\footnote{http://icvl.cs.bgu.ac.il/hyperspectral/} \cite{Arad2016Sparse},
and one remote sensing HS image, i.e.,
Urban\footnote{https://rslab.ut.ac.ir/data}. Specifically,  
  ICVL consists of 201 HS images of spatial dimensions $1392\times1300$ and spectral dimension 31 covering the wavelength in the range of 400 to 700 nm, acquired by a Specim PS Kappa DX4 HS camera. We utilized 100 HS images as the training set, and the rest as the testing set.
  Urban contains $307\times307$ pixels and 210 spectral bands collected by the HYDICE hyperspectral system. This image is corrupted by \textit{real unknown noise} and widely used for real HS image denoising testing.

Following previous works \cite{Wei20213D,Cao2021Deep}, we considered two types of noise settings, i.e., the Gaussian noise and the complex noise, which were applied to ICVL dataset to simulate noisy HS images. Specifically, for the Gaussian noise,
we set various noise levels, i.e., $\sigma=30$, $50$, $70$, and ``Blind (the value of $\sigma$ is in the range of 30 to 70 but unknown)".
We generated five types of complex noises to imitate the real-world noise cases, including Non-i.i.d. Gaussian Noise, Gaussian and Stripe Noise, Gaussian and Deadline Noise, Gaussian and Impulse Noise, and Mixture Noise. 
We refer the readers to \cite{Wei20213D,Cao2021Deep} for more details about the noise settings.\\

\noindent \textbf{Implementation details.}
We implemented all the experiments with PyTorch on a PC with NVIDIA GeForce RTX 3080 GPU, Intel(R) Core(TM) i7-10700 CPU of 2.90GHz and 64-GB RAM.
We employed the ADAM optimizer \cite{kingma2014adam} with the exponential decay rates $\beta_1=0.9$ and $\beta_2=0.999$. The total training process was 25 epochs for both two kinds of noise experiments. We initialized the learning rate as $5\times10^{-4}$, which was halved every 5 epochs. We set the batch size to 4 in all experiments. The hyper-parameter $\lambda$ is set to $5\times 10^{-5}$.

\begin{table}[t]
\caption{Comparisons of \#Param, \#FLOPs, and inference time of deep learning-based HS image denoising methods on the ICVL dataset. Since T3SC \cite{Bodrito2021a} was built based on sparse coding and deep learning, we could not calculate the \#FLOPs like other pure deep learning-based methods.
}\vspace{-0.35cm}
\centering
\begin{spacing}{1.05}
\label{tab:denoising-flops-time}
\resizebox{0.48\textwidth}{!}{
\begin{tabu}{c|c|c|c|c|c|c}
\tabucline[0.5pt]{*}
\hline
Methods &~GRN~  &QRNN3D &~T3SC~ &~SST~  &SERT  &~ReS$^3$Net~ \\  \hline\hline
\#Param (M)        &1.07  &0.86   &0.83 &4.14 &1.91  &0.66  \\ 
\#FLOPs (T)        &0.22  &1.26   & -   &1.06 &0.48  &0.95  \\  
Inference time (s) &0.23  &0.52   &1.93 &2.64 &0.67  &0.77  \\  \hline
\tabucline[0.5pt]{*}
\end{tabu}}
\end{spacing}
\end{table}

\begin{figure*}[!t]
\centering
\includegraphics[width=1\linewidth]{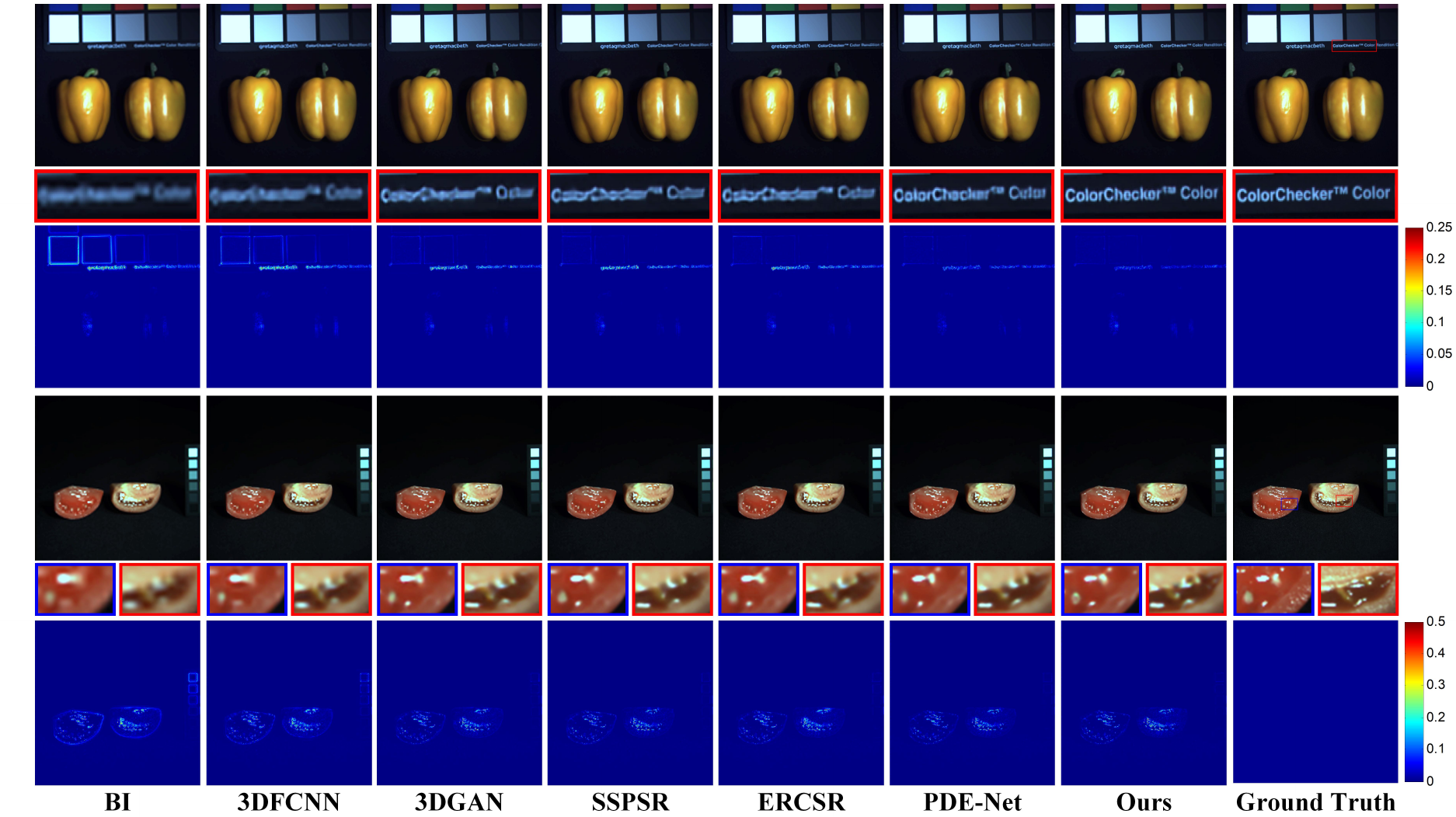}\vspace{-0.35cm}
\caption{Visual comparisons of different methods with $\alpha=4$ over CAVE dataset. For ease of comparison, we visualized the reconstructed HS images in the form of RGB images, which were generated via employing the commonly-used spectral response function of Nikon-D700 \cite{jiang2013space}. 
}
\label{fig:sr-vasual-cave}
\end{figure*}

\begin{figure*}[t]
    \centering
    \includegraphics[width=0.95\linewidth]{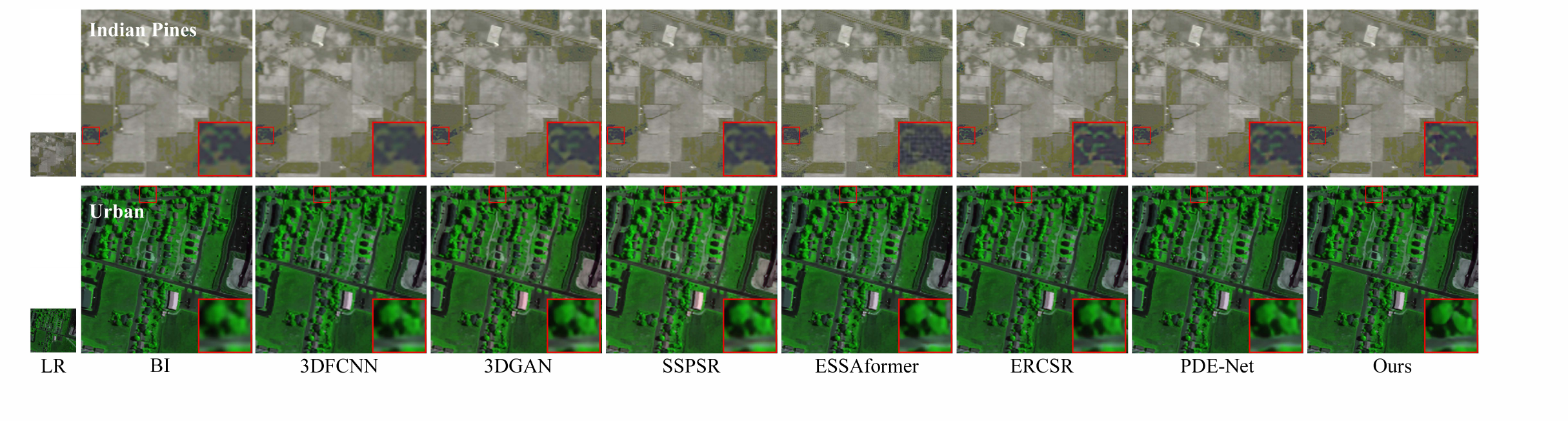}
    \caption{
    Color composite of different methods with $\alpha=4$ from Indian Pines and Urban datasets on HS image super-resolution. The test images were utilized in their original form without degradation operation. Here we selected the $16^{th}$, $15^{th}$, and $19^{th}$ bands of Indian Pines, and $3^{rd}$, $33^{rd}$, and $10^{th}$ bands of Urban, to form a pseudo-RGB image to enable the visualization.}
    \label{fig:sr_realtest}
\end{figure*}

\subsubsection{Comparison with state-of-the-art methods}

We compared the proposed denoising method with five state-of-the-art deep learning-based methods, i.e., QRNN3D \cite{Wei20213D}, GRN \cite{Cao2021Deep}, T3SC \cite{Bodrito2021a}, SST \cite{li2023spatial}, and SERT \cite{li2023spectral}  and five representative non-learning-based methods, including LRMR \cite{Zhang2014Hyperspectral}, ITSReg \cite{Xie2016Multispectral}, NMoG \cite{Chen2018Denoising}, LRTDTV \cite{Wang2018Hyperspectral}, NG-Meet \cite{he2022non}. Note that for  QRNN3D \cite{Wei20213D}, GRN \cite{Cao2021Deep}, and our method, only a single network was trained for each noise setting, i.e.,  a single network for handling noisy HS images with different  Gaussian noise levels (or complex noises);
while for T3SC \cite{Bodrito2021a}, a network was trained for each case of the two types of noise settings, i.e., nine networks for the four Gaussian noises and five complex noises. In other words, the comparison setting is more favorable for T3SC.\\

\noindent \textbf{Quantitative comparison.}
We adopted three commonly used quantitative metrics to evaluate the quality of the denoised HS images, i.e., Mean Peak Signal-to-Noise Ratio (MPSNR), Mean Structural Similarity Index (MSSIM) \cite{wang2004image}, and Spectral Angle Mapper (SAM) \cite{Yuhas1992Discrimination}.
Tables \ref{tab:icvlgaussresults} and \ref{tab:icvlcomplexresults} show the quantitative results of different methods, where it can be observed that 
our method almost achieves the best performance in terms of all three metrics under all noise scenarios.
Besides, the impressive performance of our method under  all the complex noise scenarios demonstrates  that it has better resilience on severely corrupted HS images. \\

\noindent \textbf{Visual comparison.}
Fig. \ref{fig:icvl-complex} shows the visual comparisons of denoising results by different methods on ICVL,  where we can observe that the denoised images by our method are cleaner and retain the original high-frequency details better. \\

\noindent
\textbf{Real HS denoising.}
Fig. \ref{fig:urban} provides the visual comparisons of denoised HS images by different methods on the Urban dataset, a real-world noisy HS image, where we can see that most of the compared methods fail to remove the unknown noise completely.
By contrast, our method successfully tackles this unknown noise and produces a clearer and visually pleasing image. \\

\noindent
\textbf{Computational efficiency.}
We compared the number of network parameters (\#Param), floating point of operations (\#FLOPs), and inference time of deep learning-based methods in Table \ref{tab:denoising-flops-time}, where it can be observed that our method consumes fewer network parameters and has comparable \#FLOPs, and inference time, compared with state-of-the-art methods, demonstrating that the excellent performance of our method does not come at the cost of a larger capacity and higher computational complexity but is credited to the proposed ReS$^3$-ConvSet and DA-Reg.

\begin{table}[t]
\caption{Quantitative comparisons of different methods over the CAVE dataset. The best and second-best results are highlighted in bold and underlined, respectively.``$\uparrow$" (resp. ``$\downarrow$") means the larger (resp. smaller), the better.}
\vspace{-0.2cm}
\centering
\label{tab:srcaveresults}
\resizebox{0.48\textwidth}{!}{
\begin{tabu}{c|c|c|c|c|c}
\tabucline[1pt]{*}
Methods                                  &Scale  &\#Params &MPSNR$\uparrow$  &MSSIM$\uparrow$  &SAM$\downarrow$       \\  \hline\hline  
BI                                       &4    &-        &36.533    &0.9479   &0.0738        \\ 
3DFCNN\cite{Mei2017Hyperspectral}        &4    &0.04M    &38.061    &0.9565   &0.0682        \\
3DGAN\cite{Li2020Hyperspectral}          &4    &0.59M    &39.947    &0.9645   &0.0646        \\
SSPSR\cite{Jiang2020Learning}            &4    &26.08M   &40.104    &0.9645   &0.0632        \\
ESSAformer\cite{zhang2023essaformer}     &4    &11.19M   &40.122    &0.9640   &0.0641        \\
ERCSR\cite{Li2021Exploring}              &4    &1.59M    &40.701    &\underline{0.9662}   &0.0609        \\
PDE-Net \cite{hou2022deep}  &4    &2.30M    &\underline{41.236}    &\textbf{0.9672}      &\underline{0.0603}  \\ \hline
Ours             &4    &2.11M    &\textbf{41.443} &\textbf{0.9672}   &\textbf{0.0595}    \\ \hline\hline
BI                                       &8    &-        &32.283    &0.8993   &0.0944        \\ 
3DFCNN\cite{Mei2017Hyperspectral}        &8    &0.04M    &33.194    &0.9131   &0.0876        \\
3DGAN\cite{Li2020Hyperspectral}          &8    &0.66M    &34.930    &0.9293   &0.0853        \\
SSPSR\cite{Jiang2020Learning}            &8    &28.44M   &34.992    &0.9273   &0.0817        \\
ESSAformer\cite{zhang2023essaformer}     &8    &13.70M   &35.381    &0.9292   &0.0816        \\
ERCSR\cite{Li2021Exploring}              &8    &2.38M    &35.519    &0.9338   &0.0785        \\
PDE-Net \cite{hou2022deep}      &8    &2.33M   &\underline{36.021}  &\underline{0.9363}     &\underline{0.0752}       \\ \hline
Ours                 &8    &2.31M   &\textbf{36.257}     &\textbf{0.9377}  &\textbf{0.0740}       \\
\tabucline[1pt]{*}
\end{tabu}}
\end{table}

\subsection{Evaluation on HS Image Spatial SR}

\subsubsection{Experiment settings}

\textbf{Datasets.}
We employed the commonly-used HS image benchmark dataset for evaluation, i.e., CAVE\footnote{http://www.cs.columbia.edu/CAVE/databases/} \cite{Yasuma2010CAVE},
which contains 32 HS images of spatial dimensions $512\times512$ and spectral dimension 31 covering the wavelength in the range of 400 to 700 nm, collected by a generalized assorted pixel camera. We randomly selected 20 HS images for training, and the remaining 12 HS images for testing.
Besides, we utilized two real-world HS images, i.e., Urban and Indian Pines, for real HS image super-resolution experiments.

\noindent \textbf{Implementation details.}
We employed the ADAM optimizer \cite{kingma2014adam} with the exponential decay rates $\beta_1=0.9$ and $\beta_2=0.999$.
We initialized the learning rate as $5\times10^{-4}$, which was halved every 25 epochs. We set the batch size to 4. The value of $\lambda$ is set to $1\times 10^{-7}$.
The total training process contained 50 warm-ups and 50 training epochs.
During the warm-up phase, we trained our model by only minimizing the $\ell_1$ loss, i.e., $\lambda=0$.

\begin{table}[t]
\caption{Comparisons of the computational efficiency of different HS spatial SR methods on the CAVE dataset.}\vspace{-0.35cm}
\centering
\label{tab:sr-flops-times}
\resizebox{0.49\textwidth}{!}{
\begin{tabu}{c|c|c|c|c|c|c}  \tabucline[1.5pt]{*}

Methods &Scale
&Inference time &\#FLOPs &Scale  &Inference time &\#FLOPs   \\ \hline
3DFCNN\cite{Mei2017Hyperspectral}  &4  &0.088s  &0.321T      &8   &0.080s   &0.321T    \\
3DGAN\cite{Li2020Hyperspectral}    &4  &0.369s  &1.300T      &8   &0.332s   &1.233T    \\
SSPSR\cite{Jiang2020Learning}      &4  &0.429s  &3.029T      &8   &0.302s   &1.818T    \\
ESSAformer\cite{zhang2023essaformer} &4 &0.609s  &3.168T      &8   &0.607s   &3.127T   \\
ERCSR\cite{Li2021Exploring}        &4  &0.380s  &4.463T      &8   &0.283s   &10.429T   \\
PDE-Net \cite{hou2022deep}         &4  &0.526s  &6.375T      &8   &0.207s   &3.020T    \\ \hline
Ours                               &4  &0.463s  &1.917T      &8   &0.195s   &2.752T    \\
\tabucline[1.5pt]{*}
\end{tabu}}
\end{table}

\begin{table}[t]
\caption{Quantitative performance of different classification methods in terms of OA, AA, and Kappa, as well as the accuracies for each class on the Indian Pines dataset. The best results are highlighted in bold.}\vspace{-0.35cm}
\centering
\label{tab:cls-quantt-ip}
\resizebox{0.48\textwidth}{!}{
\begin{tabu}{c|c|c|c|c}  
\tabucline[1.5pt]{*}
Class No.  &3-D CNN\cite{hamida20183} &FuNet-C\cite{hong2021graph} &SpectralFormer\cite{hong2022spectralformer} &Ours      \\  \hline \hline  
1       &\textbf{89.67}  &67.19            &68.64            &84.10            \\
2       &82.02           &78.70            &87.50            &\textbf{90.56}   \\
3       &86.96           &\textbf{99.46}   &85.33            &98.91            \\
4       &96.64           &95.75            &95.75            &\textbf{98.88}   \\
5       &91.82           &93.69            &85.80            &\textbf{98.42}   \\
6       &98.17           &\textbf{99.32}   &97.49            &98.86            \\
7       &80.39           &79.19            &\textbf{85.29}   &78.87            \\
8       &61.49           &69.77            &76.92            &\textbf{88.62}   \\
9       &80.85           &70.57            &78.90            &\textbf{87.06}   \\
10      &\textbf{100.00} &99.38            &99.38            &\textbf{100.00}  \\
11      &93.41           &89.79            &93.25            &\textbf{97.43}   \\
12      &96.96           &94.24            &81.82            &\textbf{97.58}   \\
13      &97.78           &\textbf{100.00}  &\textbf{100.00}  &\textbf{100.00}  \\
14      &89.74           &87.18            &71.79            &\textbf{100.00}  \\
15      &\textbf{100.00} &\textbf{100.00}  &\textbf{100.00}  &\textbf{100.00}  \\
16      &\textbf{100.00} &\textbf{100.00}  &\textbf{100.00}  &\textbf{100.00}  \\ \hline \hline
OA      &82.38           &80.06            &82.88            &\textbf{90.70}   \\
AA      &90.37           &89.01            &87.99            &\textbf{94.96}   \\
Kappa   &0.8004          &0.7736           &0.8047           &\textbf{0.8934}  \\ \hline
\tabucline[1.5pt]{*}
\end{tabu}}
\end{table}

\begin{table}[t]
\caption{Quantitative performance of different classification methods in terms of OA, AA, and Kappa, as well as the accuracies for each class on the Pavia University dataset. The best results are highlighted in bold.}\vspace{-0.35cm}
\centering
\label{tab:cls-quantt-pu}
\resizebox{0.48\textwidth}{!}{
\begin{tabu}{c|c|c|c|c}  
\tabucline[1.5pt]{*}
Class No.  &3-D CNN\cite{hamida20183} &FuNet-C\cite{hong2021graph} &SpectralFormer\cite{hong2022spectralformer}   &Ours      \\  \hline \hline  
1            &79.71          &80.08           &82.73           &\textbf{83.42}   \\
2            &\textbf{95.69} &94.28           &94.03           &95.31            \\
3            &80.39          &67.66           &73.66           &\textbf{85.12}   \\
4            &\textbf{98.52} &97.80           &93.75           &96.77            \\
5            &99.82          &99.19           &99.28           &\textbf{100.0}   \\
6            &85.89          &91.49           &\textbf{90.75}  &88.25            \\
7            &84.91          &85.93           &87.56           &\textbf{93.37}   \\
8            &97.24          &96.64           &95.81           &\textbf{98.54}   \\
9            &94.88          &\textbf{95.22}  &94.21           &94.84            \\ \hline \hline
OA           &91.51          &90.92           &91.07           &\textbf{92.62}   \\
AA           &90.70          &89.81           &90.20           &\textbf{92.85}   \\
Kappa        &0.8858         &0.8787          &0.8805          &\textbf{0.9010}  \\ \hline
\tabucline[1.5pt]{*}
\end{tabu}}
\end{table}

\begin{table}[t]
\caption{\textcolor{black}{Results of HS image classification on the \textit{WHU-OHS} dataset.}} \vspace{-0.3cm}
\centering
\label{tab:cls-whu}
\resizebox{0.32\textwidth}{!}{
\begin{tabu}{c|c|c|c}  %
\tabucline[1pt]{*}
Methods         &~~OA~~      &~~Kappa~~   &~~mIoU~~    \\  \hline\hline  
3-D CNN \cite{chen2016deep}  &65.19   &60.45   &40.48  \\
3-D CNN + Ours               &66.89   &62.26   &42.09  \\ 
A$^2$S$^2$K-ResNet \cite{roy2020attention}  &73.60 &70.05  &45.08   \\
FreeNet \cite{zheng2020fpga} &77.81 &74.92  &52.47     \\ 
FreeNet + Ours               &79.43 &76.77  &54.51     \\ \hline
\tabucline[1pt]{*}
\end{tabu}}
\end{table}

\begin{table}[t]
\caption{Results of the ablation study towards different rank upper bounds of our ReS$^3$-ConvSet over HS image super-resolution with $\alpha=4$ over CAVE dataset.}\vspace{-0.35cm}
\centering
\label{tab:diff-rub}
\resizebox{0.48\textwidth}{!}{
\begin{tabu}{c|c|c|c|c|c|c}  
\tabucline[1.5pt]{*}
\hline
Methods  &Rank upper bound &Kernel  &\# Params &MPSNR$\uparrow$ &MSSIM$\uparrow$ &SAM$\downarrow$   \\   \hline\hline
3-D   &1$M$   &3-D  &3.59 M   &40.98   &0.9664  &0.0602    \\
ReS$^3$-ConvSet  &3$M$   &2-D  &3.73 M   &41.28   &0.9669  &0.0597  \\
ReS$^3$-ConvSet  &3$M$   &1-D  &2.11 M   &41.35   &0.9671  &0.0596  \\
ReS$^3$-ConvSet  &6$M$   &1-D  &3.06 M   &41.29   &0.9671  &0.0596  \\
ReS$^3$-ConvSet  &9$M$   &1-D  &4.00 M   &41.32   &0.9672  &0.0592  \\ \hline
\tabucline[1.5pt]{*}
\end{tabu}}
\end{table}

\begin{table}[t]
\caption{
Results of the ablation study towards our ReS$^3$-ConvSet using different kernel sizes over HS image super-resolution with $\alpha=4$ on the CAVE dataset.
}
\vspace{-3mm}
\centering
\label{tab:ab-kernelsize-sr-cave}
\resizebox{0.5\textwidth}{!}{
\setlength{\tabcolsep}{2mm}{
\begin{tabu}{c|c|c|c|c|c}  
\tabucline[1.5pt]{*}
Method &Kernel size &\#Params (M) &MPSNR$\uparrow$ &MSSIM$\uparrow$ &SAM$\downarrow$ \\ \hline \hline
3-D conv         &3 &3.586  &40.98   &0.9664  &0.0602       \\
ReS$^3$-ConvSet  &3 &2.110  &41.35   &0.9671  &0.0596       \\ \hline
3-D conv         &5 &12.417 &40.75   &0.9658  &0.0614       \\
ReS$^3$-ConvSet  &5 &2.651  &41.40   &0.9675  &0.0590       \\ \hline
3-D conv         &7 &32.061 &40.73   &0.9662  &0.0605       \\
ReS$^3$-ConvSet  &7 &3.192  &41.48   &0.9679  &0.0587       \\ \hline
\tabucline[1.5pt]{*}
\end{tabu}}}
\end{table}

\subsubsection{Comparison with state-of-the-art methods}

We compared our HS image spatial super-resolution method with seven state-of-the-art deep learning-based methods, i.e., 3DFCNN \cite{Mei2017Hyperspectral}, 3DGAN \cite{Li2020Hyperspectral}, SSPSR \cite{Jiang2020Learning}, ERCSR \cite{Li2021Exploring}, , PDE-Net \cite{hou2022deep}, and ESSAformer\cite{zhang2023essaformer}. We also provided the results of bi-cubic interpolation (BI) as a baseline.\\

\noindent \textbf{Quantitative comparison.}
Table \ref{tab:srcaveresults} shows the quantitative results of different methods, where it can be observed that our method consistently achieves the best values in terms of all the three metrics under $4\times$ and $8\times$ spatial super-resolution tasks, validating the superiority of our proposed method. Besides, the network size of our method, validating that the satisfactory performance of our method is not achieved by piling up larger capacity but credited to elegant feature extraction technique. \\  

\noindent \textbf{Visual comparison.} Fig. \ref{fig:sr-vasual-cave} visually compares the results of different methods, where it can be observed that the super-resolved images by our method present sharper and clearer textures and are closer to the ground truth ones, which further demonstrates its advantage.\\

\begin{table*}[t]
\caption{
Results of the ablation study towards the effect of $\lambda$ on different HS image-based tasks. The best results are highlighted in bold.}
\vspace{-3.5mm}
\centering
\label{tab:ab-lamda}
\resizebox{0.95\textwidth}{!}{
\setlength{\tabcolsep}{2mm}{
\begin{tabu}{c|c|c|c|c|c|c|c|c|c|c|c}
\tabucline[1.2pt]{*}
\multicolumn{4}{c|}{HS image denoising ($\sigma=70$)}
&\multicolumn{4}{c|}{HS image SR (4$\times$)}
&\multicolumn{4}{c}{~~~HS image classification (Pavia University)~~~}\\ \hline \hline
$\lambda~(\times 10^{-6})$  &MPSNR$\uparrow$ &MSSIM$\uparrow$ &SAM$\downarrow$
&$\lambda~(\times 10^{-7})$  &MPSNR$\uparrow$ &MSSIM$\uparrow$ &SAM$\downarrow$
&$\lambda~(\times 10^{-4})$  &~~~OA$\uparrow$~~~ &~~~AA$\uparrow$~~ &Kappa$\uparrow$ \\ \hline
 1 &39.96  &0.9491  &0.0588
&1 &\textbf{41.44} &\textbf{0.9672} &0.0595
&1 &91.30  &92.92   &0.8845   \\
10 &39.99 &0.9491 &0.0584
&2 &41.40 &0.9671 &0.0596
&3 &\textbf{92.62}  &92.85 &\textbf{0.9010}   \\
50 &\textbf{40.01} &\textbf{0.9492}
&\textbf{0.0582}   &5 &41.38 &0.9670 &\textbf{0.0594}
&5 &91.58  &91.44  &0.8867   \\
100 &39.95 &0.9491 &0.0588
&10 &41.35 &0.9670 &0.0601
&10 &91.45 &\textbf{92.93}  &0.8863   \\  \hline

\tabucline[1.2pt]{*}
\end{tabu}}}
\end{table*}
\begin{figure}[t]
\centering
\includegraphics[width=1\linewidth]{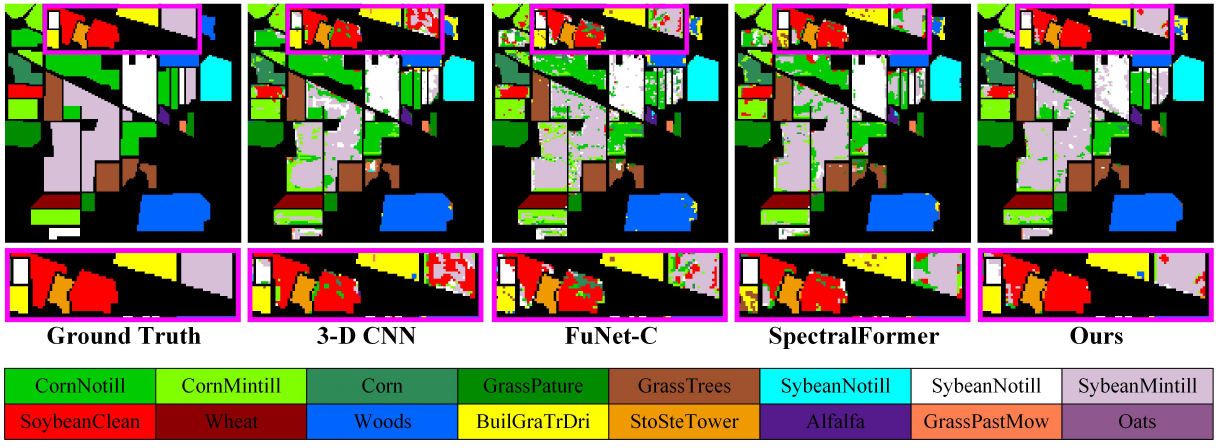}\vspace{-0.35cm}
\caption{Visual comparisons of the classification maps obtained by different methods on the Indian Pines dataset. Selected regions have been zoomed in for better comparison.
}
\label{fig:cls-vasual-ip}
\end{figure}

\begin{figure}[t]
\centering
\includegraphics[width=1\linewidth]{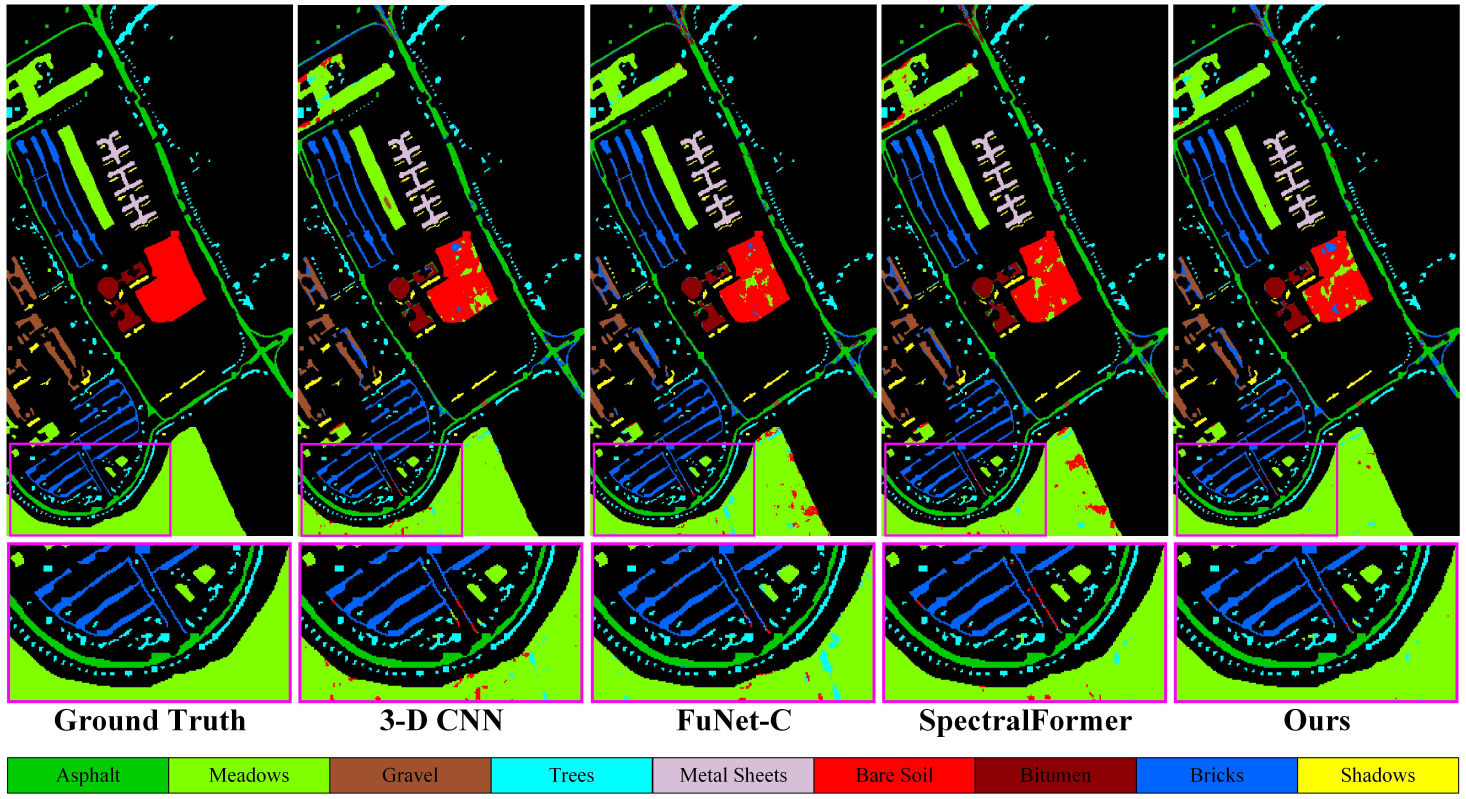}\vspace{-0.35cm}
\caption{Visual comparisons of the classification maps obtained by different methods on the Pavia University dataset. Selected regions have been zoomed in for better comparison.
}
\label{fig:cls-vasual-pu}
\end{figure}

\noindent
\textbf{Real HS spatial super-resolution.}
We also conducted real-world tests on the Indian Pines and Urban datasets to evaluate the performance of our HS image super-resolution technique. Fig. \ref{fig:sr_realtest} offers a visual comparison of the results from various HS image super-resolution methods. Inspection of these comparisons clearly shows that our proposed method exhibits superior performance in real image super-resolution tasks. \\

\noindent \textbf{Computational efficiency.}
Table \ref{tab:sr-flops-times} compares the computational efficiency of different approaches measured with the \#FLOPs and inference time, where it can be observed that the proposed method consumes much fewer \#FLOPs and fewer inference times than most compared methods.

\subsection{Evaluation on HS Image Classification}

\subsubsection{Experiment settings}

\textbf{Datasets.}
We employed the two commonly-used HS image classification benchmarks for evaluation, i.e., Indian Pines\footnote{https://rslab.ut.ac.ir/data}, and Pavia University\footnote{http://www.ehu.eus/ccwintco/index.php/Hyperspectral\_Remote\_\\Sensing\_Scenes/}. Specifically, 
  Indian Pines is of spatial dimensions $145\times145$ and spectral dimension 220 covering a wavelength in the range of 400 to 2500 nm, collected by the Airborne Visible/Infrared Imaging Spectrometer (AVIRIS) sensor. There are 16 mainly investigated categories in this studied scene. Following \cite{hong2021graph,hong2022spectralformer}, we employed 695 samples for training and 9671 samples for testing.
  Pavia University was gathered by the ROSIS sensor, consisting of $610\times 340$ pixels and $103$ spectral bands from the range of 430 to 860 nm. There are 9 land cover categories. The number of samples utilized for training and testing is 3921 and 40002, respectively.\\

\noindent \textbf{Implementation details.}
We employed the ADAM optimizer \cite{kingma2014adam} with the exponential decay rates $\beta_1=0.9$ and $\beta_2=0.999$.
The total training process contained 1000 epochs for Indian Pines, and 160 epochs for Pavia University. We set the batch size to 64 and initialized the learning rate as $1\times10^{-3}$. We set the patch size as $7\times 7$ during training\footnote{There is potential for train-test information leakage under this setting, which can be addressed using the training methods in \cite{nalepa2019validating,zou2020spectral}. However, it is essential to emphasize that our comparisons with other methods were conducted in a fair setting, allowing us to demonstrate the advantages of our proposed method effectively.}. The value of $\lambda$ is set to $3\times 10^{-4}$ for both Indian Pines and Pavia University.

\subsubsection{Comparison with state-of-the-art methods}
We compared our method with three state-of-the-art deep learning-based HS image classification methods named 3-D CNN \cite{hamida20183}, FuNet-C \cite{hong2021graph}, and SpectralFormer \cite{hong2022spectralformer}. Following existing methods, we adopted three widely-used metrics, i.e., Overall Accuracy (OA), Average Accuracy (AA), and Kappa Coefficient, to evaluate the HS image classification performance.

As listed in Tables \ref{tab:cls-quantt-ip} and \ref{tab:cls-quantt-pu},
it can be seen that our method consistently achieves the highest accuracy in terms of OA, AA, and Kappa metrics on both Indian Pines and Pavia University,
although some of the compared methods adopt more advanced deep learning strategies, such as, the fusion of CNN and GCN, and Transformer \cite{vaswani2017attention}.
Besides, from Figs. \ref{fig:cls-vasual-ip} and \ref{fig:cls-vasual-pu}, it can be seen that our method produces more accurate classification maps,
again demonstrating its advantage.

\subsubsection{Evaluation on the large-scale benchmark}
\textcolor{black}{
We further conducted experiments to verify the effectiveness of our approach on the WHU-OHS dataset \cite{li2022whu}, which comprises 42 OHS satellite images sourced from over 40 distinct locations across China. It is partitioned into training, validation, and test sets containing 4822, 513, and 2460 sub-images, respectively, each with dimensions of $512\times512$ pixels across 32 spectral bands. We adopted three metrics, i.e., OA, Kappa, and mIoU (mean Intersection over Union), to evaluate the performance. It can be observed from Table \ref{tab:cls-whu} that the advantage of our method is still verified.
}

\begin{table*}[t]
\caption{Results of the ablation study towards different convolution manners used for HS image denoising on the ICVL dataset with the Gaussian noise ($\sigma=70$).\vspace{-0.35cm}
}
\centering
\label{tab:d-ab-comb-icvl}
\resizebox{0.9\textwidth}{!}{
\begin{tabu}{c|c|c|c|c|c|c}  %
\tabucline[1pt]{*}
Methods  &\#Params (M) &\#FLOPs (T)    &Rank upper bound     &MPSNR$\uparrow$   &MSSIM$\uparrow$   &SAM$\downarrow$    \\  \hline\hline  
3-D conv. (Fig. \ref{fig:low-rank-filters} (\textcolor{red}{a}))              &1.201    &1.562    &$M$     &39.34   &0.9430   &0.0693   \\ 
Seq. 1-D conv. (Fig. \ref{fig:low-rank-filters} (\textcolor{red}{c}))         &0.614    &0.765    &$M$     &39.42   &0.9430   &0.0658   \\
Seq. 1-D and 2-D conv. (Fig. \ref{fig:low-rank-filters} (\textcolor{red}{b})) &0.717    &0.889    &$M$     &39.58   &0.9445   &0.0652   \\
1-D + 2-D conv. (Fig. \ref{fig:low-rank-filters} (\textcolor{red}{d}))        &0.740    &1.082    &$2M$    &39.71   &0.9459   &0.0610   \\
ReS$^3$-ConvSet                                                   &0.658    &0.953    &$3M$    &39.85   &0.9483   &0.0584   \\ \hline
\tabucline[1pt]{*}
\end{tabu}}
\end{table*}

\begin{table*}[t]
\caption{Results of the ablation study towards different convolution manners used for $4\times$ HS image spatial super-resolution on  the CAVE dataset.}\vspace{-0.35cm}
\centering
\label{tab:sr-ab-comb-cave}
\resizebox{0.9\textwidth}{!}{
\begin{tabu}{c|c|c|c|c|c|c}  %
\tabucline[1pt]{*}
Methods  &\#Params (M) &\#FLOPs (T)    &Rank upper bound       &MPSNR$\uparrow$   &MSSIM$\uparrow$   &SAM$\downarrow$    \\  \hline\hline  
3-D conv. (Fig. \ref{fig:low-rank-filters} (\textcolor{red}{a}))                &3.586    &2.668    &$M$     &40.98   &0.9664  &0.0602   \\ 
Seq. 1-D conv. (Fig. \ref{fig:low-rank-filters} (\textcolor{red}{c}))           &1.987    &1.853    &$M$     &41.00   &0.9668  &0.0599   \\
Seq. 1-D and 2-D conv. (Fig. \ref{fig:low-rank-filters} (\textcolor{red}{b}))   &2.269    &1.997    &$M$     &41.10   &0.9666  &0.0598   \\
1-D + 2-D conv. (Fig. \ref{fig:low-rank-filters} (\textcolor{red}{d}))          &2.332    &2.029    &$2M$    &41.17   &0.9667  &0.0596   \\
ReS$^3$-ConvSet                                                     &2.110    &1.917    &$3M$    &41.35   &0.9671  &0.0596   \\ \hline
\tabucline[1pt]{*}
\end{tabu}}
\end{table*}

\begin{table*}[t]
\caption{Results of the ablation study towards different convolution manners 
used for HS image classification on \textit{Indian Pines}.} \vspace{-0.35cm}
\centering
\label{tab:cls-ab-comb-ip}
\resizebox{0.9\textwidth}{!}{
\begin{tabu}{c|c|c|c|c|c|c}  %
\tabucline[1pt]{*}
Methods  &\#Params (M) &\#FLOPs (G)    &Rank upper bound       &OA    &AA    &Kappa    \\  \hline\hline  
3-D conv. (Fig. \ref{fig:low-rank-filters} (\textcolor{red}{a}))                &0.395    &0.0639    &$M$     &82.38   &90.37   &0.8004   \\ 
Seq. 1-D conv. (Fig. \ref{fig:low-rank-filters} (\textcolor{red}{c}))           &0.385    &0.0406    &$M$     &82.61   &90.09   &0.8023   \\
Seq. 1-D and 2-D conv. (Fig. \ref{fig:low-rank-filters} (\textcolor{red}{b}))   &0.386    &0.0419    &$M$     &82.94   &89.51   &0.8056   \\
1-D + 2-D conv. (Fig. \ref{fig:low-rank-filters} (\textcolor{red}{d}))          &0.387    &0.0442    &$2M$    &83.83   &90.97   &0.8152   \\
ReS$^3$-ConvSet                                                     &0.386    &0.0421    &$3M$    &87.84   &94.07   &0.8615   \\ \hline
\tabucline[1pt]{*}
\end{tabu}}
\end{table*}

\begin{table*}[t]
\caption{
Results of the ablation study towards our ReS$^3$-ConvSet and DA-Reg over 1) HS image denoising on the ICVL dataset with the Gaussian noise ($\sigma=70$), 2) HS image super-resolution with $\alpha=4$ on the CAVE dataset, and 3) HS image classification on \textit{Indian Pines}. For the first two settings, we replaced ReS$^3$-ConvSet with 3-D convolution with an identical number of layers.}
\vspace{-0.35cm}
\centering
\label{tab:d-ab-loss-icvl}
\resizebox{0.95\textwidth}{!}{
\begin{tabu}{cc|ccc|ccc|ccc}  
\tabucline[1.2pt]{*}
\multirow{2}{*}{ReS$^3$-ConvSet} &\multirow{2}{*}{DA-Reg}
&\multicolumn{3}{c|}{HS image denoising}
&\multicolumn{3}{c|}{HS image SR}
&\multicolumn{3}{c}{HS image classification} \\
~ &~
&MPSNR$\uparrow$ &MSSIM$\uparrow$ &SAM$\downarrow$
&MPSNR$\uparrow$ &MSSIM$\uparrow$ &SAM$\downarrow$
&OA    &AA    &Kappa \\ \hline
$\times$    &$\times$       &39.34  &0.9430  &0.0693  &40.98 &0.9664 &0.0620  &82.38  &90.37 &0.8004   \\
$\times$    &\checkmark     &39.49  &0.9438  &0.0674  &41.08 &0.9664 &0.0602  &83.66  &90.36 &0.8136   \\
\checkmark  &$\times$       &39.85  &0.9483  &0.0584  &41.35 &0.9671 &0.0596  &87.84  &94.07   &0.8615   \\
\checkmark  &\checkmark     &40.01  &0.9492  &0.0582  &41.44 &0.9672 &0.0595  &90.70  &94.96 &0.8934   \\ \hline
\tabucline[1.2pt]{*}
\end{tabu}}
\end{table*}

\begin{table*}[t]
\caption{
Results of the ablation study towards the symmetry property of our ReS$^3$-ConvSet over HS image denoising on the ICVL dataset with the Gaussian noise ($\sigma=70$) and HS image super-resolution with $\alpha=4$ on the CAVE dataset. For the ReS$^3$-ConvSet w/o symmetry, we utilized different numbers of channels for each 1-D convolution in ReS$^3$-ConvSet, i.e., the three dimensions of the HS image are unequally essential in this setting.}
\vspace{-0.35cm}
\centering
\label{tab:ab-symm-den-cave}
\resizebox{0.8\textwidth}{!}{
\begin{tabu}{c|ccc|ccc}  
\tabucline[1.5pt]{*}
\multirow{2}{*}{Methods}
&\multicolumn{3}{c|}{HS image denoising}
&\multicolumn{3}{c}{HS image SR}  \\
~
&MPSNR$\uparrow$ &MSSIM$\uparrow$ &SAM$\downarrow$
&MPSNR$\uparrow$ &MSSIM$\uparrow$ &SAM$\downarrow$   \\  \hline  
ReS$^3$-ConvSet w/o symmetry  &39.68   &0.9471 &0.0616  &41.20 &0.9670 &0.0598  \\
ReS$^3$-ConvSet               &39.85  &0.9483  &0.0584  &41.35 &0.9671 &0.0596  \\ \hline
\tabucline[1.5pt]{*}
\end{tabu}}
\end{table*}

\subsection{Ablation Study}
\label{ab-study}

\subsubsection{ReS$^3$-ConvSet with different rank upper bounds}
We compared our ReS$^3$-ConvSet with different rank upper bounds in Table \ref{tab:diff-rub}, where it can be observed that all cases
obtain better performance than the original 3-D convolution. When the rank upper bound exceeds $3M$, it only provides a modest performance improvement. We thus adopt the ReS$^3$-ConvSet with the rank upper bound of $3M$, i.e., Fig. \ref{fig:filter-rank}(\textcolor{red}{d}), in all the remaining experiments.

\subsubsection{Various Kernel Sizes}
We conducted experimental evaluations with different kernel sizes of 5 and 7 to ascertain the effectiveness and versatility of our methodology. As listed in Table \ref{tab:ab-kernelsize-sr-cave}, our ReS$^3$-ConvSet demonstrates superior performance compared to the standard 3D convolution method while maintaining a significantly lower parameter count. For instance, when employing a kernel size of 5, our ReS$^3$-ConvSet achieves superior performance metrics (MPSNR, MSSIM, and SAM) with only 2.651M parameters. In contrast, the 3D convolution method, despite utilizing a considerably larger parameter count of 12.417M, delivers results inferior to those obtained with the smaller kernel size of 3. This discrepancy in performance suggests the possibility of overfitting within the 3D convolution method. Conversely, our approach does not suffer from such a predicament.
These results demonstrate that our proposed ReS$^3$-ConvSet maintains its effectiveness and efficiency across different kernel sizes, reaffirming its versatility as a feature representation module.

\subsubsection{Ablation Study on the Hyperparameter $\lambda$}
We quantitatively compared the performance of different HS image-based tasks with varying values of $\lambda$ in Table \ref{tab:ab-lamda}. The results indicate that setting $\lambda$ to $5\times10^{-5}$ yields optimal performance for HS image denoising. Similarly, $\lambda$ values of $1\times10^{-7}$ and $3\times10^{-4}$ produce the best results for HS image SR and HS image classification, respectively, across most evaluation metrics.

\subsubsection{Contributions of ReS$^3$-ConvSet and DA-Reg}
Table \ref{tab:d-ab-loss-icvl} lists the results of our constructed denoising, super-resolution, and classification frameworks with or without the proposed ReS$^3$-ConvSet and DA-Reg. It can be observed that both ReS$^3$-ConvSet and DA-Reg make contributions to satisfactory performance.

\subsubsection{Quantitative comparisons of various convolution manners}

For fair comparisons, we built various denoising, super-resolution, and classification methods by only replacing the 1-D convolutional kernels in our ReS$^3$ block with the variants and retaining all the other settings (e.g., connections, aggregation, etc.). Besides, we also provided the results of 3-D convolution for reference.
As listed in Tables \ref{tab:d-ab-comb-icvl}, \ref{tab:sr-ab-comb-cave}, and \ref{tab:cls-ab-comb-ip}, it can be seen that compared with the 3-D convolution, all the convolution variants show their advantages on either quantitative performance or network compactness (\#Params) and complexity (\#FLOPs).
Generally, a higher upper bound of the rank produces better performance, which is consistent with our theoretical analysis.
Particularly, our ReS$^3$-ConvSet equipped with the second-fewest number of network parameters has the highest rank upper bound during the feature extraction process and thus achieves the best quantitative performance, convincingly demonstrating its superiority and the importance of filter diversity in designing feature extraction module.

\subsubsection{Distribution of singular values of the feature matrix}
We compared the distributions of the singular values of the feature matrices learned by different convolution manners.
Specifically, we unfolded the feature embeddings from different convolution patterns (i.e., 3-D, Sequential 1-D, Sequential 1-D and 2-D, 1-D + 2-D, and the proposed ReS$^3$-ConvSet) into feature matrices. Then, we computed the singular values of these feature matrices via singular value decomposition (SVD). Note that we selected those feature embeddings from the layers at the same depth for a fair comparison. As shown in Fig. \ref{fig:d-sr-cls-ab-sv-comparison}, it can be clearly seen that the singular values of the feature matrix by our methods
decrease more slowly than those of other schemes, demonstrating that the proposed methods can avoid only a few large singular values dominating the feature space, thus promoting feature diversity.

\subsubsection{The symmetry property of our ReS$^3$-ConvSet.}
We validated the importance of the symmetry property of our ReS$^3$-ConvSet. As listed in Tables \ref{tab:ab-symm-den-cave}, it can be seen that our ReS$^3$-ConvSet equipped with the symmetry property shows its advantage in reconstruction performance.

\section{Conclusion}
\label{sec:con}

In this paper, we introduced an efficient and effective feature embedding module called ReS$^3$-ConvSet, motivated by the theoretical analysis that improving the rank of the matrix formed by unfolded convolutional kernels can promote feature diversity. ReS$^3$-ConvSet, derived from modifications to the topology of 3-D convolution, promotes the upper bound of the rank by independently applying 1-D convolution along the three dimensions of HS images side-by-side; thus, it not only effectively captures the high-dimensional spatio-spectral information of HS images but also reduces computational complexity.
Furthermore, we introduced a novel diversity-aware regularization term called DA-Reg to explicitly enhance feature diversity by regularizing the singular value distribution of unfolded feature volumes. By applying the proposed ReS$^3$-ConvSet and DA-Reg to HS image denoising, spatial super-resolution, and classification tasks, we demonstrated their significant advantages. Beyond the proposed approach itself, our theoretical formulation provides a unified explanation for the various empirically designed feature embedding modules used for volume data, thereby advancing the research field in a comprehensive manner.

\bibliographystyle{IEEEtran}
\bibliography{ref}

\begin{IEEEbiography}[{\includegraphics[width=0.9in,height=1.25in,clip,keepaspectratio]{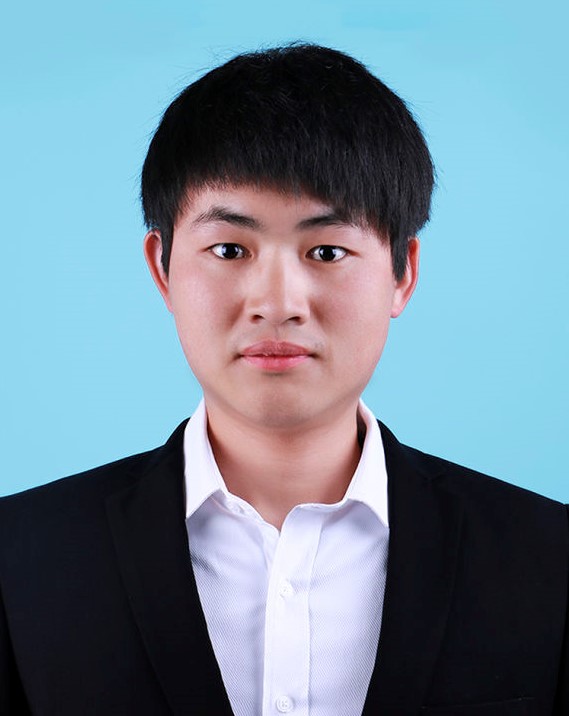}}]{Jinhui Hou} received the B.E. and M.E. degrees in communication engineering from Huaqiao University, Xiamen, China, in 2017 and 2020, respectively. He is currently pursuing the Ph.D. degree in computer science with the City University of Hong Kong. His research interests include hyperspectral image processing and deep learning.
\end{IEEEbiography}

\begin{IEEEbiography}[{\includegraphics[width=1in,height=1.25in,clip,keepaspectratio]{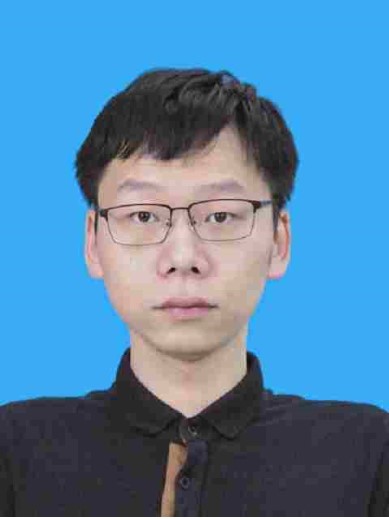}}]{Zhiyu Zhu} received the B.E. and M.E. degrees in Mechatronic Engineering, both from Harbin Institute of Technology, in 2017 and 2019, respectively. He earned his Ph.D. degree in Computer Science at the City University of Hong Kong in 2023. His research interests include computational photography and deep learning.
\end{IEEEbiography}

\begin{IEEEbiography}[{\includegraphics[width=1in,height=1.25in,clip,keepaspectratio]{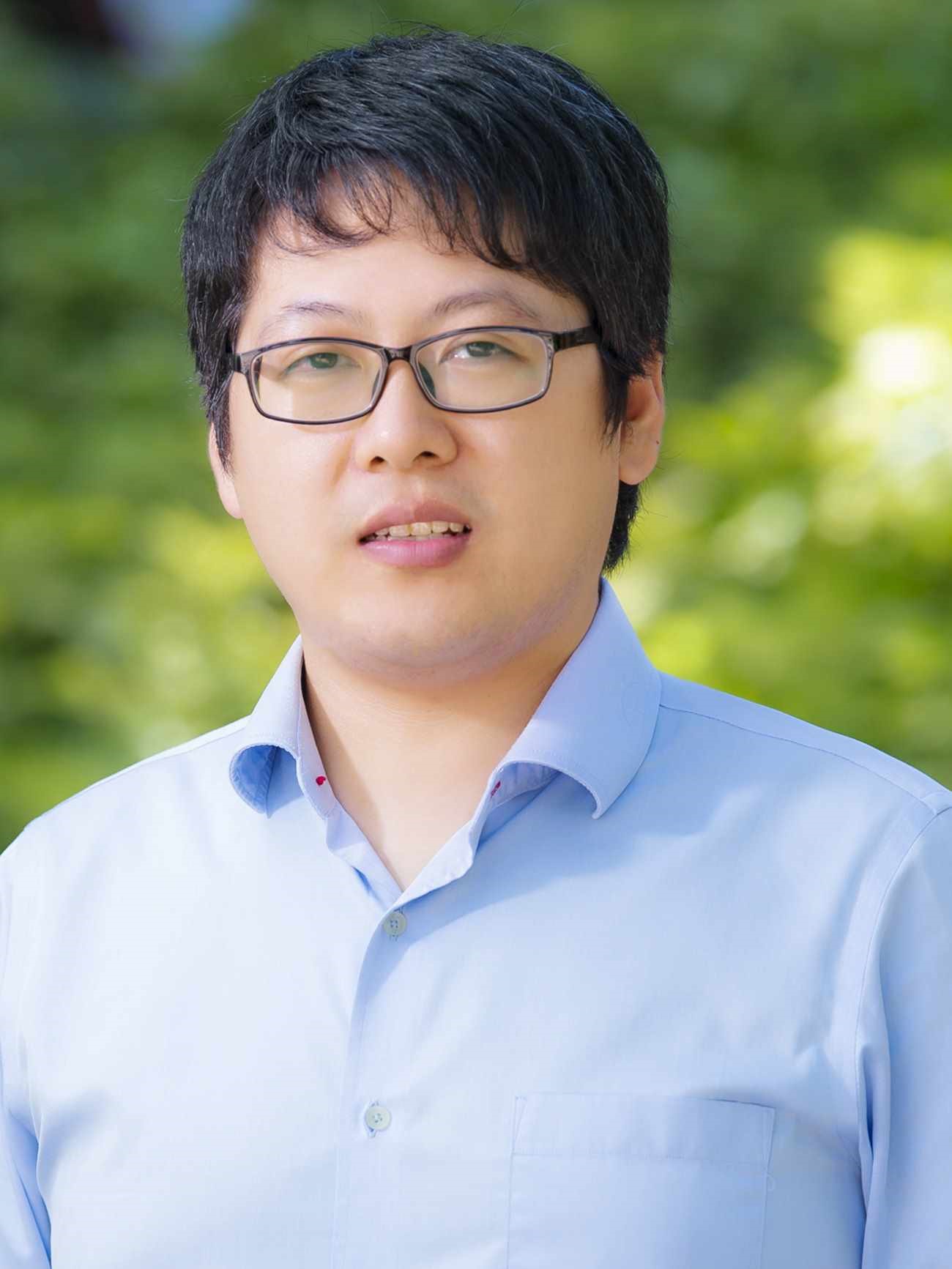}}]{Junhui Hou} is an Associate Professor with the Department of Computer Science, City University of Hong Kong. He holds a B.Eng. degree in information engineering (Talented Students Program) from the South China University of Technology, Guangzhou, China (2009), an M.Eng. degree in signal and information processing from Northwestern Polytechnical University, Xi'an, China (2012), and a Ph.D. degree from the School of Electrical and Electronic Engineering, Nanyang Technological University, Singapore (2016). His research interests are multi-dimensional visual computing.

Dr. Hou received the Early Career Award (3/381) from the Hong Kong Research Grants Council in 2018. He is an elected member of IEEE MSATC, VSPC-TC, and MMSP-TC. He is currently serving as an Associate Editor for IEEE Transactions on Visualization and Computer Graphics, IEEE Transactions on Circuits and Systems for Video Technology, IEEE Transactions on Image Processing, Signal Processing: Image Communication, and The Visual Computer.
\end{IEEEbiography}

\begin{IEEEbiography}[{\includegraphics[width=1in,height=1.5in,clip,keepaspectratio]{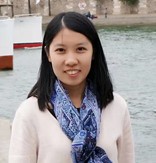}}]{Hui Liu}
 received the B.Sc. degree in communication
engineering from Central South University, Changsha,
China, the M.Eng. degree in computer science
from Nanyang Technological University, Singapore,
and the Ph.D. degree from the Department of Computer
Science, City University of Hong Kong, Hong
Kong. From 2014 to 2017, she was a Research
Associate at the Maritime Institute, Nanyang Technological
University. She is currently an Assistant
Professor with the School of Computing Information
Sciences, Saint Francis University,
Hong Kong. Her research interests include image processing and machine
learning.
\end{IEEEbiography}

\begin{IEEEbiography}[{\includegraphics[width=1in,height=1.25in,clip,keepaspectratio]{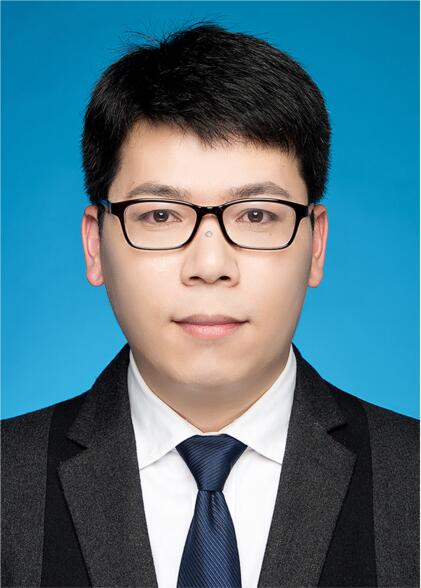}}]{Huanqiang Zeng}
received the B.S. and M.S. degrees in electrical engineering from Huaqiao University, China, and the Ph.D. degree in electrical engineering from Nanyang Technological University, Singapore.

He is currently a Full Professor at the School of Engineering and the School of Information Science and Engineering, Huaqiao University. Before that, he was a Postdoctoral Fellow at The Chinese University of Hong Kong, Hong Kong. He has published more than 180 papers in well-known journals and conferences, including three best poster/paper awards (in the International Forum of Digital TV and Multimedia Communication 2018 and the Chinese Conference on Signal Processing 2017/2019). His research interests include image processing, video coding, machine learning, and computer vision. He has also been actively serving as the General Co-Chair for IEEE International Symposium on Intelligent Signal Processing and Communication Systems 2017 (ISPACS2017), the Co-Organizer for ICME 2020 Workshop on 3D Point Cloud Processing, Analysis, Compression, and Communication, the Technical Program Co-Chair for Asia–Pacific Signal and Information Processing Association Annual Summit and Conference 2017 (APSIPA-ASC2017), the Area Chair for IEEE International Conference on Visual Communications and Image Processing (VCIP2015 and VCIP2020), and a technical program committee member for multiple flagship international conferences. He has been actively serving as an Associate Editor for IEEE Transactions on Image Processing, IEEE Transactions on Circuits and Systems for Video Technology, and IET Electronics Letters, and a Senior Area Editor for IEEE Signal Processing Letters.

\end{IEEEbiography}

\begin{IEEEbiography}[{\includegraphics[width=1in,height=1.25in,clip,keepaspectratio]{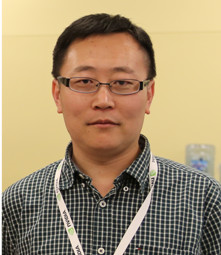}}]{Deyu Meng}received the BSc, MSc, and Ph.D. degrees from Xi'an Jiaotong University, Xi'an, China, in 2001, 2004, and 2008, respectively. He was a visiting scholar with Carnegie Mellon University, Pittsburgh, Pennsylvania, from 2012 to 2014. He is currently a professor with the School of Mathematics and Statistics, Xi'an Jiaotong University, and an adjunct professor with the Faculty of Information Technology, Macau University of Science and Technology, Taipa, Macau, China. His research interests include model-based deep learning, variational networks, and meta learning.
\end{IEEEbiography}

\end{document}